\documentclass[conference]{IEEEtran}
\IEEEoverridecommandlockouts
\usepackage{cite}
\usepackage{amsmath,amssymb,amsfonts}
\usepackage{algorithmic}
\usepackage{graphicx}
\usepackage{textcomp}
\usepackage{xcolor}
\usepackage{color}
\usepackage{url}
\usepackage{mathtools}
\usepackage{bibentry}
\usepackage{adjustbox}
\usepackage{dsfont}
\usepackage{diagbox}
\usepackage{multirow}
\usepackage{mathdots}
\usepackage{mathrsfs}
\usepackage{pifont}
\DeclareMathOperator{\E}{\mathbb{E}}
\newcommand{\dE}{\displaystyle\mathop{\mathbb{E}}}
\newcommand{\xmark}{\ding{55}}
\newcommand{\cmark}{\ding{51}}

\def\BibTeX{{\rm B\kern-.05em{\sc i\kern-.025em b}\kern-.08em
    T\kern-.1667em\lower.7ex\hbox{E}\kern-.125emX}}
\begin{document}

\title{ML4ML: Automated Invariance Testing for Machine Learning Models}

\author{\IEEEauthorblockN{Zukang Liao}
\IEEEauthorblockA{\textit{University of Oxford} \\
zukang.liao@eng.ox.ac.uk}
\and
\IEEEauthorblockN{Pengfei Zhang}
\IEEEauthorblockA{\textit{Tencent AI R\&D Centre} \\
joepfzhang@tencent.com}
\and
\IEEEauthorblockN{Min Chen}
\IEEEauthorblockA{\textit{University of Oxford} \\
min.chen@oerc.ox.ac.uk}
}
\maketitle

\begin{abstract}
In machine learning (ML) workflows, determining the invariance qualities of an ML model is a common testing procedure. Traditionally, invariance qualities are evaluated using simple formula-based scores, e.g., accuracy. In this paper, we show that testing the invariance qualities of ML models may result in complex visual patterns that cannot be classified using simple formulas.
In order to test ML models by analyzing such visual patterns automatically using other ML models, we propose a systematic framework that is applicable to a variety of invariance qualities. 
We demonstrate the effectiveness and feasibility of the framework by developing ML4ML models (assessors) for determining rotation-, brightness-, and size-variances of a collection of neural networks. Our testing results show that the trained ML4ML assessors can perform such analytical tasks with sufficient accuracy.
\end{abstract}


\section{Introduction}
\label{sec:Introduction}
In many applications, it is desirable for machine-learned models to be invariant in relation to the variation of some aspects of input data. Especially in the field of computer vision, much effort has been made to improve rotation invariance,
brightness invariance, 
size invariance, 
and other types of invariance, e.g.,
\cite{spatialinvariance1}.
Given a machine-learned model, manually determining whether the model is $\mathcal{T}$-invariant, where $\mathcal{T}$ is a type of invariance property such as rotation, can be a data-intensive and time-consuming process, which typically leads to inconsistent judgement. Hence, it is highly desirable to standardize and automate such processes.


Invariance testing has been an important part of robustness testing, and the level of invariance is traditionally measured using a relatively simple formula (e.g., \cite{goodfellow2009measuring, fuzzaugmentation}). 
Because such a formula aggregates a huge amount of measured data quickly to a single score, it cannot fully encapsulate all variance-related patterns in the measured data. Later in Section \ref{sec:Framework}, we will demonstrate that some variance patterns that should have raised some concerns may sail through a formula-based assessment.
Similar to many statistical measures (e.g., correlation index), visualization (e.g., a scatter plot) can help validate whether a statistical measure is indicative or does not meet the condition for its application (e.g., bivariate normal distribution for correlation estimation).
%

However, it is costly for human experts to inspect visual patterns in routine invariance testing in machine learning (ML) workflows, since there are different types of $\mathcal{T}$-invariance qualities and one can collect variance measurement data in different structural locations in an ML model. It is thus highly desirable to automate such visual inspection using ML.





In this paper, we address the needs (1) for improving invariance testing by allowing more detailed inspection instead of simple formula-based scores, and (2) for automating such detailed inspection instead of relying on manual visual analysis.   
We propose to use machine learning for machine learning (ML4ML), i.e., by training classifiers or regressors (ML4ML assessors) to analyse the visual patterns resulting from invariance testing. Here we use the term ``assessors" for ML4ML classifiers or regressors to avoid confusion with those ML models to be tested (the test candidates).
The main contributions of this work are:
\begin{itemize}
    \item We propose a technical framework for automated invariance testing, and we evaluate the framework for different types of invariance qualities.
    \item We use visual patterns generated at different locations using different statistical functions to show invariance qualities are more complex than formula-based scores.
    \item We collect a model repository that consists of over 600 models featuring different invariance properties for aiding the training of ML4ML assessors.
    \item We have made all the relevant models, data, and documentation available at the github \cite{liao2021ml4ml}.
    %
\end{itemize}

\section{Related work}
Existing topics of machine learning (ML) testing include
security, 
interpretability, 
fairness, 
efficiency,
and so on. Zhang et al. provided a comprehensive survey on these topics \cite{mltestsurvey}. In this section, we focus on \emph{invariance testing} and the related robustness testing.

The invariance qualities of ML models have been studied for several decades \cite{Mundy:1993:book}. One common approach to improving such quality is through augmentation
\cite{augsurvey}. 
Engstrom et al. \cite{ExploringLandscape} found that augmentation was crucial for improving the robustness of ML models since adversarial attacks may be carried out by rotating or translating inputs. Zhang \cite{translation} found that CNNs were not fully translation-invariant because MaxPooling layers were not optimal in terms of Nyquist sampling theory.
However, in previous works, invariance qualities were mostly measured by accuracy-based formulas, e.g., 
\cite{revisitingaugmentation, encodeaugmentation}.
This work extends such evaluation by considering visual patterns generated at different positions of ML models.

Compared with gradient-based methods 
(e.g., \cite{advattack, simuladvtrain}),
adversarial examples generated 
using affine transformation are more closely related to invariance testing.
The works on neuron coverage, such as DeepXplore \cite{deepxplore} and DeepGauge \cite{deepgauge} are highly representative. TensorFuzz \cite{tensorfuzz} introduced a property-based fuzzing procedure to enhance its convergence.
DLFuzz \cite{dlfuzz} exploited a coverage-guided fuzzing approach to generate transformed adversarial examples.
DeepHunter \cite{deephunter} introduced a frequency-aware seed selection strategy. 
These methods were able to find adversarial examples using affine transformations, while our framework focuses on evaluating how robust a model is under transformation(s).

Selective classification (or reject options) 
\cite{anomaloussurvey}
and out-of-distribution detection \cite{oodsurvey} are also closely related to robustness testing. 
Werpachowski et al. \cite{rejectoption} trained a reject function for each trained CNN, and they used either softmax response (confidence score) or uncertainty \cite{uncertainty} to judge if the CNN is confident enough for a given data object $x$. We adopt confidence scores as one of the testing criteria in our framework. The approach was typically for evaluating the likelihood of untrustworthy predictions (e.g., \cite{gramood}), while our work focuses on evaluating invariance qualities of ML models as a whole.

Metamorphic testing \cite{metamorphicsurvey} is also strongly related to invariance testing. Different types of mutations can be applied to permutation on input channels or the order of training/testing data \cite{metamorphic} as well as model structures \cite{munn, DeepMutation}. Both metamorphic and invariance testing can be used to prevent unwanted deployment, although they serve different purposes for different properties.

\begin{figure*}[t]
    \centering
    \includegraphics[height=79mm]{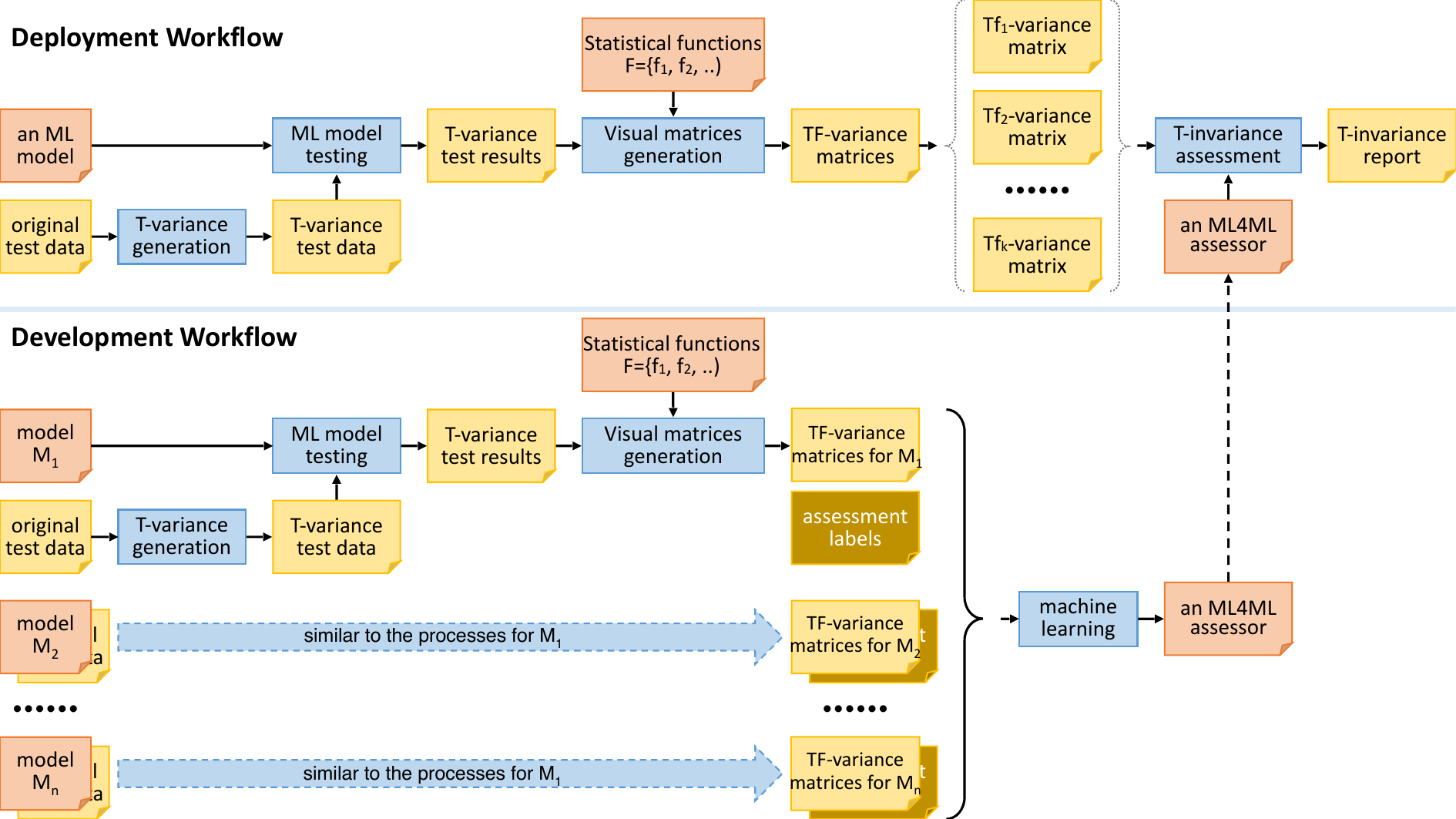}
    \caption{\textbf{Deployment workflow:} for each type of invariance $\mathcal{T}$, the original test data is first extended to invariance test data, which is then used to test the ML model concerned. The invariance test data is processed using different statistical functions (which are also referred to as \emph{modalities}), yielding different visual patterns (which are also referred to as \emph{variance matrices}).
    An ML4ML assessor (a model / an ensemble of models) analyzes the variance matrices and reports the invariance quality of the ML model concerned.
    \textbf{Development workflow:} A repository of ML models is collected. The same processes for creating variance matrices are applied to these models. The labelled variance matrices are used to train an ML4ML assessor that is to be used in the deployment workflow.}
    \label{fig:Workflow}
\end{figure*}

\section{Definitions and Motivation}
\label{sec: definition}

In machine learning, the invariance quality of a trained model $\mathbf{M}$ is typically defined as the level of consistency when the model makes predictions.
Consider a transformable attribute of input data objects, such as the rotation, brightness, or size of images. The more consistent the predictions of the model regardless of the variation of the attribute are, the better invariance quality the model has.

Let $\mathbb{D}$ be the set of all possible data objects that could be the input of model $\mathbf{M}$, which is commonly referred to as the data space of $\mathbf{M}$.
Let $x$ and $y$ be two arbitrary data objects in $\mathbb{D}$.
Let $D \subset \mathbb{D}$ be a subset where all data objects are largely similar but vary in one particular aspect, denoted as variation $\mathcal{T}$.
Consider every such subset, $D_1, D_2, \ldots$ in $\mathbb{D}$.

\noindent\textbf{Definition.} $\mathbf{M}$ is said to be $\mathcal{T}$-invariant if for any pair of data objects both belonging to the same subset $D_i \; (i=1, 2, \ldots)$, i.e., $\forall x, y \in D_i$, we have $\mathbf{M}(x) = \mathbf{M}(y)$.

When $\mathbf{M}$ is not $\mathcal{T}$-invariant, ideally we can assess the level of inconsistency (or  $\mathcal{T}$-variance) with a function $\Upsilon$:
\[
    \text{Level}_{\mathcal{T}} = \Upsilon(\mathbf{M}, D_1, D_2, \ldots)
\]
Obtaining such an ideal measurement Level$_{\mathcal{T}}$ is almost impossible because (i) it is difficult to identify all subsets $D_1, D_2, \ldots$ and obtain all data objects in such a subset; (ii) the definition of ``largely similar'' is imprecise and subjective; and (iii) there is no ground truth function for $\Upsilon$.

Hence the level of $\mathcal{T}$-variance of a model is commonly estimated by using a simplified process. One such example is to use a family of transformations $T$ to simulate variation $\mathcal{T}$ \cite{ExploringLandscape}.
One obtains a testing dataset $X \subset \mathbb{D}$, and for each data object $x_i \in X$ $(i=1,2,\ldots, m)$, one applies different transformation $t_j \in T$ $(j=1, 2, \ldots, n)$ to $x_i$ to create a set of data objects:
\[
    \tilde{D}_i = \{ d_{i,0}, d_{i,1}, \ldots, d_{i, n}\} = \{ t_0(x_i), t_1(x_i), \ldots, t_n(x_i) \}
\]
%
\noindent One can define a function $f$ to measure the inconsistency between two predictions of model $\mathbf{M}$. With $f$ one may approximate $\Upsilon$ with the following simplistic function $\Phi$:
\begin{equation}
\label{eq:RobustAcc}
    \Phi(\mathbf{M}, X) = \frac{1}{m}\sum_{i=1}^m \prod_{j=1}^{n} f \bigl( \mathbf{M}(x_i), \mathbf{M}(t_j(x_i)) \bigr) 
\end{equation}
\noindent When $f=1$, $x_i$ and $t_j(x_i)$ are consistently classified, i.e., $\mathbf{M}(x_i)=\mathbf{M}(t_j(x_i))$, and when $f=0$ the contrary. Because Gao et al. (\cite{fuzzaugmentation}) also associated consistency with correctness, i.e.,  $\mathbf{M}(x_i) = \mathbf{M}(t_j(x_i)) = \textit{correct prediction}$, they referred such a $\Phi$ to as \emph{robust accuracy}. Eq.\,\ref{eq:RobustAcc} is served as a simple baseline in this work.

As mentioned earlier, this is not a ground truth function, and $\Phi$ can be defined differently, e.g., using a mean function instead of the product in Eq.\,\ref{eq:RobustAcc}. In this work, we made use of the simulation approach based on a family of transformations $T$. While exploring more general ways of defining the function $f$, e.g., different statistical functions, we utilise ML techniques to obtain the overall function $\Phi$.
Using ML techniques to aid the evaluation of the invariance quality of ML models allows us to capture and encode complex knowledge about multivariate invariance assessment.


\section{Methodology: Testing Framework}
\label{sec:Framework}
As shown in Fig. \ref{fig:Workflow}, our ML4ML framework for invariance testing includes two workflows: a \emph{development workflow} and a \emph{deployment workflow}.
In the former workflow (lower part of Fig. \ref{fig:Workflow}), one collects a repository of models and uses them to train ML4ML assessors for invariance testing.
For each type of invariance $\mathcal{T}$, this workflow will yield one assessment model or an ensemble of models, which are to be used in the latter workflow (upper part of Fig. \ref{fig:Workflow}).

\subsection{Deployment Workflow}
\label{sec:Deployment}
\paragraph{Transformations and Sampling Variable.} Let us first describe the deployment of an ML4ML assessor for testing a type of invariance $\mathcal{T}$. Given an ML model $\mathbf{M}$ and a testing dataset $X$, we first define a \emph{family of transformation functions} $T=\{ t_0, t_1, \ldots, t_n \}$ to simulate the variations for testing $\mathcal{T}$. We use an \emph{independent variable} $V$ to characterize such variations, e.g., the rotation degree of an image for testing rotation-invariance or the average image brightness for testing brightness-invariance. When $V = 0$, the corresponding $t()$ is an identity transformation. 
We can thus order the functions in $T$ incrementally according to $V$.

We recommend to define $T$ to enable the sampling of $V$ in a fixed interval $[-\alpha, \alpha]$ with $2k+1$ instances ($k \in \mathbb{N}^+$). Hence $n=2k$ and $t_{k}() = t_{n/2}()$ is an identity transformation.
For example, to simulate rotation variations, one may create a family of transformations $T_r$ corresponding to:
\[
    V_r = \{ -90^\circ, -87^\circ, \ldots, -3^\circ, 0^\circ, 3^\circ, \ldots, 87^\circ, 90^\circ \}
\]
\noindent To simulate brightness variances, one may create a family of transformations $T_b$ corresponding to:
\[
    V_b = \{ -25\%, -24\%, \ldots, -1\%, 0\%, 1\%, \ldots, 24\%, 25\% \}
\]
\noindent Note that $T_r$ may feature other variations in addition to rotation degrees (e.g., image cropping or background filling), and $T_b$  may feature other variations in addition to brightness changes (e.g., contrast normalization or color balancing). These additional adjustments are usually there to minimize the confounding effects that may be caused by the primary variation $V_r$ or $V_b$.
If one wishes to test the invariance quality related to any of these additional variation qualities, one can specify a new type $\mathcal{T}_{new}$, define a new set of transformations $T_{new}$, and a new sampling variable $V_{new}$.

\begin{figure*}[t]
\centering
\includegraphics[height=31mm]{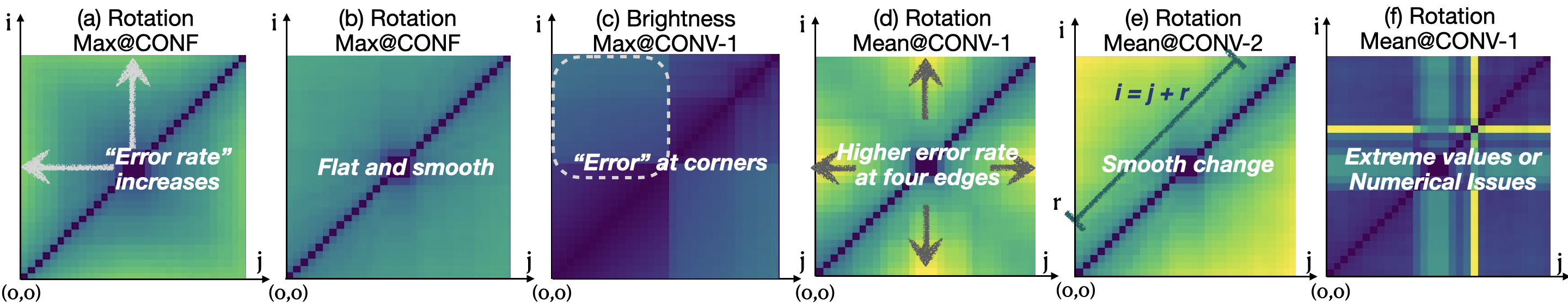}
\caption{Six examples of variance matrices for different CNNs.
All matrices are of size $31 \time 31$, and were created with a family of 31 transformations. Their testing purposes, transformations, positions, and modalities are detailed in Table \ref{table:examples}.}
\label{fig:pattern_example}
\end{figure*}

\begin{table*}[t]
\centering
\caption{Technical specifications for obtaining the six examples of variance matrices shown in Fig. \ref{fig:pattern_example}.}
\begin{adjustbox}{max width=164mm, center}
\begin{tabular}{|c|c|c|c|c|c|}
\hline
\textbf{Matrix} & \textbf{Variance} & \textbf{Testing Interval} & \textbf{Testing Position} & \textbf{Function} $\mathrm{dif}$ & \textbf{Model ID} \\ \hline
(a) & rotation   & {[}-15$^\circ$, 15$^\circ${], $\pm1^\circ$}      & confidence score (CONF) & $\max(S_x) - \max(S_y)$ & $\mathbf{M}_{93}$    \\ \hline
(b) & rotation   & {[}-15$^\circ$, 15$^\circ${], $\pm1^\circ$}      & confidence score (CONF) & $\max(S_x) - \max(S_y)$ & $\mathbf{M}_{2}$    \\ \hline
(c) & brightness & {[}-30\%, +30\%{], $\pm2\%$} & last convolutional layer (CONV-1) & $\max(S_x) - \max(S_y)$ & $\mathbf{M}_{101}$    \\ \hline
(d) & rotation   & {[}-15$^\circ$, 15$^\circ${], $\pm1^\circ$}      & last convolutional layer (CONV-1) & $\mathrm{mean}(S_x) - \mathrm{mean}(S_y)$ & $\mathbf{M}_{93}$    \\ \hline
(e) & rotation   & {[}-15$^\circ$, 15$^\circ${], $\pm1^\circ$}      & penultimate convolutional layer (CONV-2) & $\mathrm{mean}(S_x) - \mathrm{mean}(S_y)$ & $\mathbf{M}_2$    \\ \hline
(f) & rotation   & {[}-15$^\circ$, 15$^\circ${], $\pm1^\circ$}      & last convolutional layer (CONV-1) & $\mathrm{mean}(S_x) - \mathrm{mean}(S_y)$ & $\mathbf{M}_1$    \\ \hline
\end{tabular}
\end{adjustbox}
\label{table:examples}
\end{table*}

\paragraph{Modality.} As described earlier, with the defined transformations $T$, we can create a testing data set $\tilde{D}_i$ for each data object $x_i$ in the original testing dataset $X$. We can then measure the impact of the transformations $t_j$ $(j=0, 1, \ldots, n)$ upon $x_i$. Eq.\,\ref{eq:RobustAcc} used a simple function $f$ to measure the difference between the model predictions based on $x_i$ and $t_j(x_i)$ respectively.
Inspired by medical imaging \cite{modality}, we replace the simple $f$ with different statistical functions. And we also replace the ``summation on production", i.e., $\sum_{i} \prod_{j}$, operation in Eq.\,\ref{eq:RobustAcc} with a different cumulative function. We refer a combination of a statistical function and cumulative function to as a \emph{modality}. Additionally, we would like to observe not only the output of a model $\mathbf{M}$ but also the internal status of $\mathbf{M}$ in response to $x_i$ and $t_j(x_i)$, which are referred to as different \emph{modalities} at different \emph{locations}.

We thus extend the definition of $f$. Consider the data flowing from an input $x \in X$, through a model $\mathbf{M}$, to the final prediction of $\mathbf{M}$. We can measure the signals at an internal position $p$ of $\mathbf{M}$ or its output layer.
We define $f$ that maps a set of position-sensitive signals to a measurement.
Let $S$ be a set of signals at a position of $\mathbf{M}$, such as the output vector/tensor at a layer in a CNN model, a feature vector handled by a decision-tree model, or a confidence vector generated by a classifier.
$S$ is thus sensitive to the model $\mathbf{M}$, a specific position $p$, and a specific input data object $x$.
We denote this as $S(\mathbf{M}, p, x)$.

Given two different input data objects $x$ and $y$, we can obtain two signal sets $S(\mathbf{M}, p, x)$ and $S(\mathbf{M}, p, y)$. We can measure the difference between $S(\mathbf{M}, p, x)$ and $S(\mathbf{M}, p, y)$ using a difference function:
\[
    \mathrm{dif}\bigl( S(\mathbf{M}, p, x), S(\mathbf{M}, p, y) \bigr)
\]
\noindent Because $S$ is a set of signals, $\mathrm{dif}()$ is more complicated than the simple $f()$ function used in Eq.\,\ref{eq:RobustAcc}. Recall the notations of an original testing dataset $X$, a set of transformations $T$ and its corresponding sampling variable $V$. 
We can use $\mathrm{dif}()$ to measure the difference between any two derived testing data objects $t_i(x)$ and $t_j(x)$ as:
\begin{equation}
\label{eq:Dif}
    \mathrm{dif}\bigl( S(\mathbf{M}, p, t_i(x)), S(\mathbf{M}, p, t_j(x)) \bigr)
\end{equation}
\noindent Let us rewrite $S(\mathbf{M}, p, t_i(x))$ as $S_i  = \{ s_{i,1}, s_{i,2}, \ldots\}$ and $S(\mathbf{M}, p, t_j(x))$ as $S_j = \{ s_{j,1}, s_{j,2}, \ldots\}$.
The $\mathrm{dif}$ function can be as simple as $\mathrm{dif}_a = \max(S_i) - \max(S_j)$ or $\mathrm{dif}_b = \mathrm{mean}(S_i) - \mathrm{mean}(S_j)$, or as complex as a Minkowski distance metric or other difference functions.

For all $x_k \in X$, we can measure the cumulative effect as:
\[
    \delta_{i,j} = \underset{x_k \in X}{\mathrm{cumul}} \Bigl(
      \mathrm{dif}\bigl( S(\mathbf{M}, p, t_i(x_k)), S(\mathbf{M}, p, t_j(x_k)) \bigr)
    \Bigr)
\]

The functions $\mathrm{dif}()$ and $\mathrm{cumul}()$ define the essence of a mapping from a set of signals detected at a position $p$ of a model $\mathbf{M}$ to a measurement, and can thus be considered as \emph{modalities} in analogy to medical imaging 
\cite{modality}
. We refer a combination of $\mathrm{dif}()$ and $\mathrm{cumul}()$ as a \emph{modality}.

In this work, we use the same $\mathrm{cumul}$ function for all our modalities. When there is no confusion which model $\mathbf{M}$ and which testing position $p$ we are referring to, we denote Eq. \ref{eq:Dif} as $\mathrm{dif_{i,j}}()$.
The function $\mathrm{cumul}$ is thus defined as:   
\[
    \delta_{i,j} =
    \underset{x_k \in X}{\mathrm{cumul}} = \sqrt{
    \underset{x_k \in X}{\mathbb{E}} \Bigl\{ [\mathrm{dif_{i,j}}(x_k)]^2 \Bigr\} }
\]
\noindent where $\mathbb{E}$ is the expected value.
Using standard concentration inequalities, such as the Hoeffding or Bernstein bounds \cite{hdp}, we can show that with this $\mathrm{cumul}$ function, $\delta_{i,j}$ concentrates around the true expected value $\hat{\delta}_{i,j}$ unless there are extreme values or numerical issues. The detailed proof can be found in Appendix D on our github repository for this work \cite{liao2021ml4ml}. 

\begin{figure*}[t]
    \centering
    \includegraphics[height=49mm]{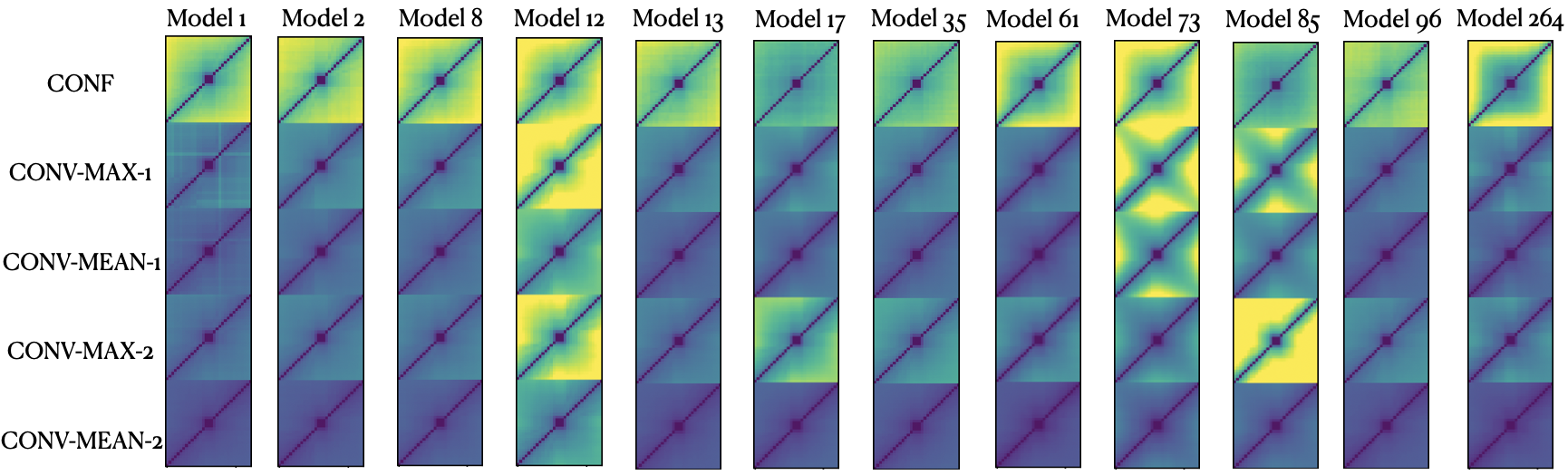}
    \caption{Five variance matrices for twelve models with almost the same robust accuracy were arbitrarily selected from the repository of 600 CNNs. All twelve models were trained on CIFAR10. The transformations simulate rotation variance, in the range of [-15$^\circ$, 15$^\circ$]$\pm1^{\circ}$. The metadata of these models can be found in Table 6
    in the Appendix C \cite{liao2021ml4ml}
    .
    From the images, we can easily observe different visual patterns. We extract 80 features from each variance matrix. Four example features are given in Table \ref{table:12measurements}.}
    \label{fig:ml4ml_exp}
\end{figure*}

\begin{table*}[t]
\centering
\caption{For each of the twelve models in Fig. \ref{fig:ml4ml_exp}, we show the measurements of four example features, together with the class labels used for training ML4ML assessors for testing rotation invariance. The bold numbers indicate some anomalous values.}
\resizebox{160mm}{!}{
\begin{tabular}{|c|c|c|c|c|c|c|c|c|c|c|c|c|}
\hline
\diagbox[width=55mm]{Measurements}{Model} & 1 & 2 & 8 & 12 & 13 & 17 & 35 & 61 & 73 & 85 & 96 & 264 \\ \hline
Max@CONF: Square Mean & \textbf{0.015} & 0.013 & 0.014 & \textbf{0.015} & 0.012 & 0.007 & 0.010 & 0.013 & \textbf{0.016} & 0.008 & 0.011 & 0.014 \\\hline

Max@CONV-1: Gradient score & 0.594 & 0.819 & 0.879 & 0.991 & 0.939 & \textbf{1.028} & 0.928 & \textbf{1.072} & \textbf{1.032} & \textbf{1.002} & 0.853 & 0.942 \\ \hline

Max@CONV-2: Discontinuity & 0.936 & 0.810 & 0.878 & \textbf{1.401} & 0.929 & \textbf{1.345} & 0.944 & \textbf{1.456} & \textbf{2.416} & \textbf{2.059} & 0.758 & \textbf{1.530} \\ \hline

Mean@CONV-1: Asymmetry & \textbf{0.047} & 0.009 & 0.008 & 0.014 & 0.007 & 0.012 & 0.008 & 0.009 & \textbf{0.041} & \textbf{0.021} & 0.008 & 0.012 \\ \hline

$\ldots$ & $\ldots$ & $\ldots$ & $\ldots$ & $\ldots$ & $\ldots$ & $\ldots$ & $\ldots$ & $\ldots$ & $\ldots$ & $\ldots$ & $\ldots$ & $\ldots$ \\ \hline

``Labels" for invariance & \xmark & \cmark & \cmark & \xmark & \cmark & \xmark & \cmark & \xmark & \xmark & \xmark & \cmark & \xmark \\ \hline
\end{tabular}}
\label{table:12measurements}
\end{table*}

\paragraph{Variance Matrix.} When we measure $\delta_{i,j}$ for all pairs of transformations in
$T=\{ t_0, t_1, \ldots, t_n \}$, we can obtain a \emph{variance matrix}:
\[
    \Delta(\mathbf{M}, T, X, p, \mathrm{dif}, \mathrm{cumul}) = \begin{bmatrix}
        \delta_{n, 0} & \delta_{n, 1} & \cdots & 0 \\
        \vdots  & \vdots & \iddots & \vdots  \\
        \delta_{1, 0} & 0 & \cdots & \delta_{1, n}\\
        0  & \delta_{0, 1} & \cdots & \delta_{0, n} \\ 
    \end{bmatrix}
\]
where the elements in the matrix are ordered according to the sampling variable $V$ horizontally and vertically. 
In Fig. \ref{fig:Workflow}, we use $f_1, f_2, \ldots$ to denote different modalities $(\mathrm{dif}, \mathrm{cumul})$, and use $F$ to denote a set of modalities.
When different modalities are used to probe different positions of a model, we can obtain a collection of variance matrices like multi-modality in medical imaging \cite{multimodality}.

The variance matrices can indeed be viewed as images. We thus intentionally place the matrix element $\delta_{0,0}$ at the bottom-left of the matrix to be consistent with the directions of coordinate axes. Fig. \ref{fig:pattern_example} shows six examples of such images, which are referred to as \emph{variance matrices}.


\paragraph{Image Analysis.} One can observe a variety of imagery patterns from the variance matrices as illustrated in Fig. \ref{fig:pattern_example}.
Each $\delta_{i,j}$ value is encoded using a colormap with dark blue for little cumulative difference and yellow for greater difference.
The typical patterns that one may observe include:
\begin{itemize}
    \item \emph{Incremental Direction}. By definition, the matrix cells on the diagonal lines, $\delta_{i,i}$, should exhibit no difference. As both axes of the matrix correspond to the range $[-\alpha, \alpha]$ of an independent variable $V$ that characterizes a type of variance $\mathcal{T}$, one may anticipate that the amount of inconsistency will normally increase from the matrix center $(0, 0)$ towards the cells furthest away from the diagonal lines, i.e., $(-\alpha, \alpha)$ and  $(\alpha, -\alpha)$. Examples (a), (b), and (e) in Fig. \ref{fig:pattern_example} exhibit the expected incremental direction, while (c), (d), and (f) reveal less expected patterns.
    \item \emph{Incremental Rate}. The colors depict the rate of increasing difference effectively. Among the three expected patterns, (b) shows a slow increment and (e) shows a rapid increment, while (a) is somewhere in the middle.
    \item \emph{Transitional Smoothness}. While the transitional patterns in (a), (b), (d), and (e) are smooth, those in (c) and (f) show some irregularities that attract viewers' attention.  
    \item \emph{Irregularity: Sub-domain Partition}. There can be many types of irregularities. Example (c) shows one type of irregularity, which may suggest that the model under testing has different behaviors in different sub-domains of $[-\alpha, \alpha] \times [-\alpha, \alpha]$. In this case, positive and negative variations of $V$ impact the model differently.
    \item \emph{Irregularity: dot-patterns}. Example (d) shows relatively fast increments (from the center of the matrix) towards the centers of the four edges, i.e., $(0, \pm\alpha)$ and $(\pm\alpha, 0)$. This indicates that the inconsistency between $0$ and $\pm\alpha$ is greater than the inconsistency between $+\alpha$ and $-\alpha$.
    %
    \item \emph{Irregularity: Abrupt Transition}. Example (f) shows one weak and one strong abrupt pattern. The strong pattern (shown in yellow) suggests that the testing may be affected by a specific transformation with an adversary effect, or the model being tested cannot handle certain input conditions very well. The weak pattern (shown in cyan) suggests the possibility of combined irregularities of sub-domain partition and dot-patterns.        
\end{itemize}

All the different visual patterns mentioned above show that a simple formula-based score does not suffice to approximate invariance qualities. In Fig. \ref{fig:ml4ml_exp} we also show twelve models that have similar robust accuracy on the interval of [-15$^\circ$, 15$^\circ$] ($\sim$66.5\%$\pm$1.5\%), however, their variance matrices are all different. Therefore, we propose to use the aforementioned visual patterns as a general guideline to provide invariance annotations.

\paragraph{Invariance Data Labels.} The judgments about different patterns can vary from person to person, especially when such images are unfamiliar to many potential human annotators. Therefore we hire three professionals in ML/DL, and we provide them with all the metadata (details can be found in the Appendix B \cite{liao2021ml4ml}
) and the visual patterns shown in Fig. \ref{fig:ml4ml_exp}.

In many fields, e.g., facial action units \cite{actionunits}, natural language processing (NLP) \cite{imtiaz2018sentiment}, segmentation \cite{noisylabel_segmentation}, etc, annotations are not always accurate or reliable. When there is no ``clean ground truth" for evaluation, we adopt majority votes as the ``pseudo ground truth" which was proven to be superior than individual annotators by \cite{spam_filter} using differential comparison. As suggested by \cite{irr}, when providing new research approaches/annotations, we use two inter-rater reliability (IRR) measures, i.e., Cohen’s kappa \cite{cohen1960coefficient} and Fleiss’ kappa \cite{fleiss1971measuring}, to evaluate how much confidence we can have on our invariance labels.

We consider that the level of $\mathcal{T}$-invariance of ML models should be assessed as a generic quality measurement, while the requirements for invariance quality may be applicant-dependent. For example, given two ML models for object detection, the annotations for $\mathcal{T}$-invariance should be a score consistently regardless of their intended applications. Meanwhile, different applications, e.g., detecting vegetables in photos and detecting humans in CCTV imagery, likely require different levels of rotation- or brightness-invariance.

\subsection{Development Workflow}
In order to avoid manually evaluating invariance qualities, we proposed ML4ML where we train an ML4ML assessor to automatically perform $\mathcal{T}$-invariance assessment. For a given simulation of $\mathcal{T}$, to train an ML4ML assessor, we first need to collect ``training data", which in our case are the variance matrices captured using a group of modalities $F$ at different locations. In Fig. \ref{fig:Workflow}, variance matrices are labelled with the prefix ``TF-'' to emphasize the dependency of the transformations on the testing data (i.e., the simulation) and modalities used in the workflow.

Because the ML4ML assessor for $\mathcal{T}$-invariance is application-independent, we can use different ML models and different versions of them to generate training data, i.e., variance matrices, as long as each variance matrix is clearly identified with its \emph{modality} and \emph{position}.
An ML4ML assessor is normally trained with data sharing the same modality and position properties, though future ML techniques may enable learning from heterogeneous training data.
In most serious machine learning workflows, e.g., in academic research and industrial development, many models or many versions of a model are being trained and tested.
Hence, gathering a repository of variance matrices from different workflows collectively will be attainable in the longer term.
For this work, we demonstrate the feasibility of the proposed framework using a small repository of 600 CNNs.

\paragraph{Decision Tree Learning.}
We use techniques in the decision tree (DT) family to train our ML4ML assessors, because (1) we have a relatively smaller model repository, for which DT techniques are more suitable; (2) invariance quality evaluation needs to be explainable and DT models are more transparent and understandable; and (3) annotations of invariance qualities may include noise due to subjectivity, bagging methods often achieve superior results \cite{noisylabel_survey, noisylabel_intro}. 

Given a family of transformations $T = \{t_0, t_1, \ldots t_n \}$, a set of $m$ modalities $F = \{ f_1, f_2, \ldots, f_m\}$ and a set of $l$ positions $P = \{ p_1, p_2, \ldots, p_l\}$, there are $m \cdot l$ variance matrices $\Delta$, each of $(n+1)^2$ dimensions.
The goal of an ML4ML assessor $\mathbf{Q}$ is to take the $m \cdot l$ variance matrices as input and predict the level of invariance quality.

We use binary labels (0: invariant, 1: variant) to annotate models in a model repository $\mathcal{M} = \{\mathbf{M}_1, \mathbf{M}_2, \ldots \}$. Since $\mathbf{Q}$ is trained with such labelled data, it produces a binary class label $\in \{0, 1\}$ normally, or a real value $\in [0, 1]$ if it is an ensemble or regression model.
Note that the formula in Eq.\,\ref{eq:RobustAcc} by \cite{fuzzaugmentation} defines 1 as fully invariant. 
%
%

\paragraph{Features in Variance Matrices.}
When the DT techniques and regression method are applied to variance matrices, it is necessary to extract features from these matrices. 
Given an $(n+1) \times (n+1)$ variance matrix $\Delta = [\delta_{i,j} ]$ $(i,j \in [0, n])$, a feature is a variable that characterize one aspect of the matrix. Similar to many ML solutions in medical image analysis, we adopt the methodology of ``a bag of features'' by defining a collection of features for characterizing different aspects of a variance matrix, including but not limited to gradient, asymmetry, continuity, etc (details in the Appendix A \cite{liao2021ml4ml}) 
). 

Each feature extraction algorithm is thus a function $h: \mathbb{R}^{(n+1) \times (n+1)} \rightarrow \mathbb{R}$.
In this work, we have defined a total of 16 features (see the Appendix A \cite{liao2021ml4ml}
) for each variance matrix.
The $m \cdot l$ variance matrices thus yield $16 m \cdot l$ features. For example, Fig. \ref{fig:ml4ml_exp} shows five variance matrices for each of the 12 models in the model repository.
Two modalities (labelled as Max and Mean) and three positions (CONF, CONV-1, and CONV-2) are shown in the figure, from which we can observe different visual features. Metadata about these 12 models can be found in Table 6
in Appendix C \cite{liao2021ml4ml}
. Such metadata is not used in training any ML4ML assessor.

Table \ref{table:12measurements} shows four types of feature measurements obtained from some of these images, together with the labels of invariance quality. In each row, there are one or a few numbers (in bold) that are higher than most of the others, indicating the ability of a feature to differentiate some anomalies. 

\section{Experimental Results and Analysis}
\label{sec:experiment}

\paragraph{Model Repository.}
To build a model repository, we trained 600 CNNs under different settings. The types of variations of the settings include:
(a) CNN structures (VGG13bn and CNN5);
(b) Training datasets (CIFAR10 and MNIST);
(c) Normalization of pixel values ($[-0.5, 0.5]$, $[0, 1]$);
(d) Learning rate ($e^{-2}, e^{-3}, e^{-4}, e^{-5}$);
(e) Number of epochs ($\in [1, 2000]$); Batch size (4, 8, 1024, and 16384);
(f) Optimizers (Adam or SGD);
(g) Anomalies (\emph{none}, data leakage);
(h) Training with adversarial attacks (\emph{none}, FGSM);
(i) Hardware (CPU, GPU);
(j) Added variance (\emph{none}, rotation, scaling, brightness);
(k) Transformation ranges (\emph{none}, different $[-\alpha, \alpha]$); and
(l) some minor variations.

The majority of the models were trained using Tesla V100 32GB. When they were used for training or testing ML4ML assessors, the experiments were run on a single laptop (Apple M1 Chip) without GPU.
In Appendix B \cite{liao2021ml4ml}
, we provide further details on the training parameters, while in Appendix C \cite{liao2021ml4ml}
we describe the structure of CNN5 and the metadata of the models trained with this structure.

\paragraph{Invariance labels.}
To assess the quality of the data labels used for training, we measured the inter-rater reliability (IRR) of the acquired data labels.
As shown in Table \ref{table:irr}, the IRR scores (i.e., the Cohen's and Feliss' kappa scores) are around 0.76 to 0.82, which are similar to the quality of the annotations measured in a few NLP applications \cite{irr}.

\begin{table}[t]
\caption{IRR: Cohen's and Fleiss' kappa scores}
\centering
\begin{adjustbox}{width=83mm, center}
\begin{tabular}{|c|c|c|c|c|c|}
\hline
Cohen’s & Coder 1 & Coder 2 & Coder 3 & Pseudo GT & Fleiss’ \\ \hline
Coder 1 & 1       & 0.818   & 0.801   & 0.927               &         \\ \cline{1-5}
Coder 2 & 0.818   & 1       & 0.763   & 0.890               & 0.793   \\ \cline{1-5}
Coder 3 & 0.801   & 0.763   & 1       & 0.871               &         \\ \hline
\end{tabular}
\end{adjustbox}
\label{table:irr}
\end{table}

\begin{table*}[t]
\caption{Comparing the performance (accuracy) of ML4ML assessors and a numerical assessor in four development workflows.}
\begin{adjustbox}{width=164mm, center}
\begin{tabular}{|c|c|c|c|c|}
\hline
\multirow{2}{*}{\diagbox[width=50mm]{Classifier/Regressor}{Transformation}}& \multicolumn{2}{c|}{\emph{Rotation with}}      & \emph{Brightness with}        & \emph{Scaling with} \\ \cline{2-5} 
                  & \emph{VGG13bn on CIFAR10} & \emph{CNN5 on MNIST} & \emph{VGG13bn on CIFAR10} & \emph{VGG13bn on CIFAR10} \\ \hline
Decision Tree     & 84.73\%$\pm$4.03\% & 82.93\%$\pm$3.73\% & 84.01\%$\pm$3.34\% & 91.33\%$\pm$5.19\% \\ \hline
Random Forest     & \textbf{91.31\%$\pm$2.02\%} & \textbf{87.66\%$\pm$1.02\%} & \textbf{91.40\%$\pm$2.12\%} & \textbf{94.66\%$\pm$1.13\%} \\ \hline
AdaBoost          & 89.90\%$\pm$4.62\% & 85.46\%$\pm$2.26\% & 89.93\%$\pm$4.20\% & 93.86\%$\pm$3.33\% \\ \hline
Linear Regression & 75.46\%$\pm$3.13\% & 68.26\%$\pm$2.93\% & 72.70\%$\pm$6.77\% & 78.80\%$\pm$3.20\% \\ \hline
\emph{baseline} \cite{fuzzaugmentation} & \emph{76.67}\% & \emph{83.33}\% & \emph{79.33}\% & \emph{78.67}\% \\ \hline
\multicolumn{5}{l}{\emph{All ML4ML results are averaged on ten repeated experiments. For each experiment, the standard 3-fold cross-evaluation is used.}}
\end{tabular}
\end{adjustbox}
\label{table:ml4ml_results}
\end{table*}

\paragraph{Training and Testing ML4ML Assessors.}
We trained three ML4ML assessors using the techniques in the decision tree family, including decision trees, random forests, and AdaBoost. We also trained an ML4ML assessor using linear regression. We used the scikit-learn libraries \cite{scikit-learn} for the implementation.
We set all the hyper-parameters of the assessors, e.g., depth, to default values without tuning. For regressors, we set the threshold to 0.5 without further tuning.

In addition, we implemented a numerical assessor using Eq.\,\ref{eq:RobustAcc} as the baseline for comparison. We used greedy search to find the threshold on the training data.

To train each ML4ML assessor, we selected a relatively balanced subset of ``variant" and ``invariant" models according to the type of invariance $\mathcal{T}$ that the assessor was targeted for.
This selection process resulted in four different development workflows denoted in Table \ref{table:ml4ml_results} as:

\indent
$W_1$: \emph{Rotation with VGG13bn (models) on CIFAR10},\\
\indent
$W_2$: \emph{Rotation with CNN5 (models) on MNIST},\\
\indent
$W_3$: \emph{Brightness with VGG13bn (models) on CIFAR10},\\
\indent
$W_4$: \emph{Scaling with VGG13bn (models) on CIFAR10}.

Details on the setup of training and testing the ML4ML assessors of the four workflows $W_1\sim W_4$ can be found in Appendix C \cite{liao2021ml4ml}
and an ablation study in Appendix E \cite{liao2021ml4ml}
.

\paragraph{Variance Matrices} To create variance matrices for the models, we used four families of 31 transformations. The ranges of these transformations and the fixed intervals are:

\indent
$W_1, W_2$: $[-15^{\circ}, 15^{\circ}]\;\pm1^{\circ}$,\\
\indent
$W_3, W_4$: $[-30\%, +30\%]$ or [$\times$70\%, $\times$130\%]$\;\pm2\%$.

For the 600 CNNs, we select three interested positions: at confidence score level (CONF), and at the last two convolutional layers (CONV-1 and CONV-2). And we used two $\mathrm{dif}$ functions: $\max(S_x) - \max(S_y)$ and $\mathrm{mean}(S_x) - \mathrm{mean}(S_y)$. The former was applied to CONF, and both were applied to CONV-1 and CONV-2. 
Details on the selection of testing locations can be found in Appendix B \cite{liao2021ml4ml}
.

\textbf{Feature Extraction.} For each variance matrix, we generated 16 feature measurements (a full list in Appendix A \cite{liao2021ml4ml}
), resulting in a $5\times16=80$ dimensional \emph{feature vector}. 
Table 5
in Appendix A \cite{liao2021ml4ml}
shows that some feature measurements at the CONF level have a strong correlation with the robust accuracy defined in Eq.\,\ref{eq:RobustAcc}.
However, as shown in Fig. \ref{fig:ml4ml_exp}, it does not suffice to use robust accuracy only to approximate invariance qualities. Therefore we extracted multiple feature measurements at different locations to facilitate the training of ML4ML assessor(s), and thus replace the simple formula-based accuracy score defined in Eq. \ref{eq:RobustAcc}.

\paragraph{Result Analysis.}
In each of four development workflows, $W_1 \sim W_4$, we trained four ML4ML assessors, together with a numerical assessor as the baseline for comparison. The performances of these assessors are shown in Table \ref{table:ml4ml_results}, from which we can make the following observations:
\begin{itemize}
    \item The three DT assessors performed noticeably better than linear regression and the numerical assessor (Eq.\,\ref{eq:RobustAcc}). The only exception is perhaps with workflow \emph{Rotation with CNN5 on MNIST}, where the numerical assessor performed similarly to the three DT assessors.
    \item Random forest consistently performs better than AdaBoost. This is aligned with the observation in \cite{noisylabel_survey, noisylabel_intro, spam_filter} where bagging methods were shown to counteract noisy annotations, while AdaBoost tends to place larger weights on noisy or mislabelled instances.
    \item The results of \emph{CNN5 on MNIST} are not as good as other workflows, except for the numerical assessor. Further investigation will be necessary to see if the main factors reside with the CNN5 structure, the MNIST dataset, or the training of the ML4ML assessors. 
\end{itemize}

The results in Table \ref{table:ml4ml_results} allow us to draw the following conclusions: (i) Our feature selection was effective, otherwise, the DT assessors would not perform notably better than others; (ii) The notion of variance matrices may be generally applicable to different types of variance, otherwise, the results could vary more dramatically among different variance testing; (ii) The size of our model-repository is relatively small, and the statistics (e.g., entropy) is less accurate. Although random forest and boosting can help, it is desirable to continue expanding the model repository.

\section{Conclusions}

In this work, we propose a novel framework for evaluating the invariance qualities of ML models. This framework can be used to automate the testing procedure, replace labour-intensive analysis, and prevent unwanted deployment of the ML models.

We use visualisation to show that simple formula-based scores, e.g., robust accuracy, does not suffice to approximate invariance qualities. The proposed framework enables us to evaluate invariance qualities from different perspectives, i.e., different modalities, testing positions and visual patterns.

We also show the inter-rater reliability scores ($\sim$0.76-0.82) of invariance annotations, which would be 
 considered between ``Good'' and ``Excellent'' in
typical NLP applications \cite{irr}.

Our experimentation on three different types of invariance qualities with 600 CNN models has confirmed the usability and feasibility of the proposed technical framework, as the three ML4ML assessors based on DT techniques can achieve around $83 \sim 95\%$ accuracy, noticeably better than linear regression and the robust accuracy.



\bibliography{aitest}

\begin{thebibliography}{10}

\bibitem{anomaloussurvey}
Saikiran Bulusu, Bhavya Kailkhura, Bo~Li, Pramod~K Varshney, and Dawn Song.
\newblock Anomalous example detection in deep learning: A survey.
\newblock {\em IEEE Access}, 8:132330--132347, 2020.

\bibitem{cohen1960coefficient}
Jacob Cohen.
\newblock A coefficient of agreement for nominal scales.
\newblock {\em Educational and psychological measurement}, 20(1):37--46, 1960.

\bibitem{spam_filter}
Gordon~V Cormack and Aleksander Kolcz.
\newblock Spam filter evaluation with imprecise ground truth.
\newblock In {\em Proceedings of the 32nd international ACM SIGIR conference on
  Research and development in information retrieval}, pages 604--611, 2009.

\bibitem{spatialinvariance1}
Eric Crawford and Joelle Pineau.
\newblock Spatially invariant unsupervised object detection with convolutional
  neural networks.
\newblock In {\em Proceedings of the AAAI Conference on Artificial
  Intelligence}, pages 3412--3420, 2019.

\bibitem{noisylabel_segmentation}
Rodrigo~Caye Daudt, Adrien Chan-Hon-Tong, Bertrand Le~Saux, and Alexandre
  Boulch.
\newblock Learning to understand earth observation images with weak and
  unreliable ground truth.
\newblock In {\em IGARSS IEEE International Geoscience and Remote Sensing
  Symposium}, pages 5602--5605. IEEE, 2019.

\bibitem{imagenet}
Jia Deng, Wei Dong, Richard Socher, Li-Jia Li, Kai Li, and Li~Fei-Fei.
\newblock Imagenet: A large-scale hierarchical image database.
\newblock In {\em IEEE conference on computer vision and pattern recognition},
  pages 248--255. Ieee, 2009.

\bibitem{metamorphic}
Anurag Dwarakanath, Manish Ahuja, Samarth Sikand, Raghotham~M Rao,
  RP~Jagadeesh~Chandra Bose, Neville Dubash, and Sanjay Podder.
\newblock Identifying implementation bugs in machine learning based image
  classifiers using metamorphic testing.
\newblock In {\em Proceedings of the 27th ACM SIGSOFT International Symposium
  on Software Testing and Analysis}, pages 118--128, 2018.

\bibitem{irr}
N~El~Dehaibi and EF~MacDonald.
\newblock Investigating inter-rater reliability of qualitative text annotations
  in machine learning datasets.
\newblock In {\em Proceedings of the Design Society: DESIGN Conference},
  volume~1, pages 21--30. Cambridge University Press, 2020.

\bibitem{ExploringLandscape}
Logan Engstrom, Brandon Tran, Dimitris Tsipras, Ludwig Schmidt, and Aleksander
  Madry.
\newblock Exploring the landscape of spatial robustness.
\newblock In {\em International Conference on Machine Learning}, pages
  1802--1811. PMLR, 2019.

\bibitem{fleiss1971measuring}
Joseph~L Fleiss.
\newblock Measuring nominal scale agreement among many raters.
\newblock {\em Psychological bulletin}, 76(5):378, 1971.

\bibitem{noisylabel_intro}
Beno{\^\i}t Fr{\'e}nay, Ata Kab{\'a}n, et~al.
\newblock A comprehensive introduction to label noise.
\newblock In {\em ESANN}. Citeseer, 2014.

\bibitem{noisylabel_survey}
Beno{\^\i}t Fr{\'e}nay and Michel Verleysen.
\newblock Classification in the presence of label noise: a survey.
\newblock {\em IEEE transactions on neural networks and learning systems},
  25(5):845--869, 2013.

\bibitem{uncertainty}
Yarin Gal and Zoubin Ghahramani.
\newblock Dropout as a bayesian approximation: Representing model uncertainty
  in deep learning.
\newblock In {\em international conference on machine learning}, pages
  1050--1059. PMLR, 2016.

\bibitem{fuzzaugmentation}
Xiang Gao, Ripon~K Saha, Mukul~R Prasad, and Abhik Roychoudhury.
\newblock Fuzz testing based data augmentation to improve robustness of deep
  neural networks.
\newblock In {\em IEEE/ACM 42nd International Conference on Software
  Engineering (ICSE)}, pages 1147--1158. IEEE, 2020.

\bibitem{goodfellow2009measuring}
Ian Goodfellow, Honglak Lee, Quoc Le, Andrew Saxe, and Andrew Ng.
\newblock Measuring invariances in deep networks.
\newblock {\em Advances in neural information processing systems}, 22:646--654,
  2009.

\bibitem{dlfuzz}
Jianmin Guo, Yu~Jiang, Yue Zhao, Quan Chen, and Jiaguang Sun.
\newblock Dlfuzz: differential fuzzing testing of deep learning systems.
\newblock In {\em Proceedings of the 26th ACM Joint Meeting on European
  Software Engineering Conference and Symposium on the Foundations of Software
  Engineering}, pages 739--743, 2018.

\bibitem{imtiaz2018sentiment}
Nasif Imtiaz, Justin Middleton, Peter Girouard, and Emerson Murphy-Hill.
\newblock Sentiment and politeness analysis tools on developer discussions are
  unreliable, but so are people.
\newblock In {\em IEEE/ACM 3rd International Workshop on Emotion Awareness in
  Software Engineering (SEmotion)}, pages 55--61. IEEE, 2018.

\bibitem{adam}
Diederik~P Kingma and Jimmy Ba.
\newblock Adam: A method for stochastic optimization.
\newblock {\em arXiv:1412.6980}, 2014.

\bibitem{multimodality}
Ashnil Kumar, Jinman Kim, Weidong Cai, Michael Fulham, and Dagan Feng.
\newblock Content-based medical image retrieval: a survey of applications to
  multidimensional and multimodality data.
\newblock {\em Journal of digital imaging}, 26(6):1025--1039, 2013.

\bibitem{liao2021ml4ml}
Zukang Liao.
\newblock {ML4ML} invariance testing.
\newblock \url{https://github.com/Zukang-Liao/ML4ML-invariance-testing}.

\bibitem{simuladvtrain}
Zukang Liao.
\newblock Simultaneous adversarial training-learn from others’ mistakes.
\newblock In {\em 14th IEEE International Conference on Automatic Face \&
  Gesture Recognition}, pages 1--7. IEEE, 2019.

\bibitem{actionunits}
Yen~Khye Lim, Zukang Liao, Stavros Petridis, and Maja Pantic.
\newblock Transfer learning for action unit recognition.
\newblock {\em arXiv:1807.07556}, 2018.

\bibitem{deepgauge}
Lei Ma, Felix Juefei-Xu, Fuyuan Zhang, Jiyuan Sun, Minhui Xue, Bo~Li, Chunyang
  Chen, Ting Su, Li~Li, Yang Liu, et~al.
\newblock Deepgauge: Multi-granularity testing criteria for deep learning
  systems.
\newblock In {\em Proceedings of the 33rd ACM/IEEE International Conference on
  Automated Software Engineering}, pages 120--131, 2018.

\bibitem{DeepMutation}
Lei Ma, Fuyuan Zhang, Jiyuan Sun, Minhui Xue, Bo~Li, Felix Juefei-Xu, Chao Xie,
  Li~Li, Yang Liu, Jianjun Zhao, et~al.
\newblock Deepmutation: Mutation testing of deep learning systems.
\newblock In {\em IEEE 29th International Symposium on Software Reliability
  Engineering (ISSRE)}, pages 100--111. IEEE, 2018.

\bibitem{Mundy:1993:book}
Joseph~L. Mundy, Andrew Zisserman, and David Forsyth, editors.
\newblock {\em Proc. Second Joint European-US Workshop on Applications of
  Invariance in Computer Vision}, volume 825 of {\em Lecture Notes in Computer
  Science}.
\newblock Springer, 1993.

\bibitem{modality}
Ghulam Murtaza, Liyana Shuib, Ainuddin~Wahid Abdul~Wahab, Ghulam Mujtaba,
  Henry~Friday Nweke, Mohammed~Ali Al-garadi, Fariha Zulfiqar, Ghulam Raza, and
  Nor~Aniza Azmi.
\newblock Deep learning-based breast cancer classification through medical
  imaging modalities: state of the art and research challenges.
\newblock {\em Artificial Intelligence Review}, 53(3):1655--1720, 2020.

\bibitem{tensorfuzz}
Augustus Odena, Catherine Olsson, David Andersen, and Ian Goodfellow.
\newblock Tensorfuzz: Debugging neural networks with coverage-guided fuzzing.
\newblock In {\em International Conference on Machine Learning}, pages
  4901--4911. PMLR, 2019.

\bibitem{pytorch}
Adam Paszke, Sam Gross, Soumith Chintala, Gregory Chanan, Edward Yang, Zachary
  DeVito, Zeming Lin, Alban Desmaison, Luca Antiga, and Adam Lerer.
\newblock Automatic differentiation in pytorch.
\newblock \url{https://openreview.net/pdf?id=BJJsrmfCZ}, 2017.

\bibitem{scikit-learn}
F.~Pedregosa, G.~Varoquaux, A.~Gramfort, V.~Michel, B.~Thirion, O.~Grisel,
  M.~Blondel, P.~Prettenhofer, R.~Weiss, V.~Dubourg, J.~Vanderplas, A.~Passos,
  D.~Cournapeau, M.~Brucher, M.~Perrot, and E.~Duchesnay.
\newblock Scikit-learn: Machine learning in {P}ython.
\newblock {\em Journal of Machine Learning Research}, 12:2825--2830, 2011.

\bibitem{deepxplore}
Kexin Pei, Yinzhi Cao, Junfeng Yang, and Suman Jana.
\newblock Deepxplore: Automated whitebox testing of deep learning systems.
\newblock In {\em proceedings of the 26th Symposium on Operating Systems
  Principles}, pages 1--18, 2017.

\bibitem{revisitingaugmentation}
Facundo Quiroga, Franco Ronchetti, Laura Lanzarini, and Aurelio~F Bariviera.
\newblock Revisiting data augmentation for rotational invariance in
  convolutional neural networks.
\newblock In {\em International Conference on Modelling and Simulation in
  Management Sciences}, pages 127--141. Springer, 2018.

\bibitem{gramood}
Chandramouli~Shama Sastry and Sageev Oore.
\newblock Detecting out-of-distribution examples with gram matrices.
\newblock In {\em International Conference on Machine Learning}, pages
  8491--8501. PMLR, 2020.

\bibitem{metamorphicsurvey}
Sergio Segura, Dave Towey, Zhi~Quan Zhou, and Tsong~Yueh Chen.
\newblock Metamorphic testing: Testing the untestable.
\newblock {\em IEEE Software}, 37(3):46--53, 2018.

\bibitem{munn}
Weijun Shen, Jun Wan, and Zhenyu Chen.
\newblock Munn: Mutation analysis of neural networks.
\newblock In {\em IEEE International Conference on Software Quality,
  Reliability and Security Companion (QRS-C)}, pages 108--115. IEEE, 2018.

\bibitem{augsurvey}
Connor Shorten and Taghi~M Khoshgoftaar.
\newblock A survey on image data augmentation for deep learning.
\newblock {\em Journal of Big Data}, 6:1--48, 2019.

\bibitem{advattack}
Christian Szegedy, Wojciech Zaremba, Ilya Sutskever, Joan Bruna, Dumitru Erhan,
  Ian Goodfellow, and Rob Fergus.
\newblock Intriguing properties of neural networks.
\newblock {\em arXiv:1312.6199}, 2013.

\bibitem{hdp}
Roman Vershynin.
\newblock {\em High-dimensional probability: An introduction with applications
  in data science}, volume~47.
\newblock Cambridge university press, 2018.

\bibitem{rejectoption}
Roman Werpachowski, Andr{\'a}s Gy{\"o}rgy, and Csaba Szepesv{\'a}ri.
\newblock Detecting overfitting via adversarial examples.
\newblock {\em arXiv:1903.02380}, 2019.

\bibitem{toxic}
Ellery Wulczyn, Nithum Thain, and Lucas Dixon.
\newblock Ex machina: Personal attacks seen at scale.
\newblock In {\em Proceedings of the 26th international conference on world
  wide web}, pages 1391--1399, 2017.

\bibitem{deephunter}
Xiaofei Xie, Lei Ma, Felix Juefei-Xu, Minhui Xue, Hongxu Chen, Yang Liu,
  Jianjun Zhao, Bo~Li, Jianxiong Yin, and Simon See.
\newblock Deephunter: a coverage-guided fuzz testing framework for deep neural
  networks.
\newblock In {\em Proceedings of the 28th ACM SIGSOFT International Symposium
  on Software Testing and Analysis}, pages 146--157, 2019.

\bibitem{encodeaugmentation}
Eddie Yan and Yanping Huang.
\newblock Do cnns encode data augmentations?
\newblock {\em arXiv:2003.08773}, 2020.

\bibitem{oodsurvey}
Jingkang Yang, Kaiyang Zhou, Yixuan Li, and Ziwei Liu.
\newblock Generalized out-of-distribution detection: A survey.
\newblock {\em arXiv:2110.11334}, 2021.

\bibitem{mltestsurvey}
Jie~M Zhang, Mark Harman, Lei Ma, and Yang Liu.
\newblock Machine learning testing: Survey, landscapes and horizons.
\newblock {\em IEEE Transactions on Software Engineering}, 2020.

\bibitem{translation}
Richard Zhang.
\newblock Making convolutional networks shift-invariant again.
\newblock In {\em International Conference on Machine Learning}, pages
  7324--7334. PMLR, 2019.

\end{thebibliography}
\bibliographystyle{plain}

\section*{Appendices}

\section{Feature Measurements}
\label{sec:feature_measurements}

To provide abstraction of each variance matrix, we consider 16 types of feature measurements. All ML models involve abstraction. Feature engineering is an effective way for introducing human analytical knowledge to ML models, while data labelling provides spontaneous human knowledge to models.
Both can have biases and be costly. The former is usually provided by one or a few experts, while the latter is usually provided by many less-skilled data annotators. Learning only from labelled knowledge usually requires very large training datasets. In ML, feature engineering can reduce such demand, allowing models to be trained with relatively small training datasets.

\paragraph{1. Squared Value Mean (all elements).} The mean of the squared values of all elements in the variance matrix.
\[
    \phi_{\text{svm}} = \frac{1}{2(n+1)^2} \sum\nolimits^n_{i=0} \sum\nolimits^n_{j=0} \delta^2_{i,j}
\]

\paragraph{2. Mean (``meaningful'' elements).} The mean value of all ``meaningful'' elements in the variance matrix (i.e., the upper left part) after excluding the duplication and the diagonal line (self-comparison).
\[
    \phi_{\text{mean}} = \frac{2}{n(n+1)} \sum\nolimits^{n}_{i=1} \sum\nolimits^{i-1}_{j=0} \delta_{i,j}
\]

\paragraph{3. Standard Deviation.} The standard deviation of all ``meaningful'' elements considered in (2) in the variance matrix (i.e., the upper left part).
\[
    \phi_{\text{std}}^2 = \frac{2}{n(n+1)} \sum\nolimits^{n}_{i=1} \sum\nolimits^{i-1}_{j=0} (\delta_{i,j}-\phi_{\text{mean}})^2
\]

\paragraph{4. Amount of Significant Variance.}
Given a threshold value $\tau$, any value $\delta_{i,j}$ (pairwise variance) is considered to be undesirable or significant if $\delta_{i, j} > \tau$.
In this work, we set $\tau=0.15$.
\begin{align*}
    \nu_{i,j} =& \begin{cases}
        1 & \text{if } \delta_{i, j} > \tau\\
        0 & \text{if } \delta_{i, j} \leq \tau
    \end{cases}\\
    \phi_{\text{asv}} =& \frac{2}{n(n+1)} \sum\nolimits^{n}_{i=1} \sum\nolimits^{i-1}_{j=0} \nu_{i,j}
\end{align*}

\paragraph{5. Sensitivity.} Sensitivity level to the size of testing dataset. Normally the value of $\delta_{i,j}$, i.e.,
\[
    \delta_{i,j} =
    \underset{x_k \in X}{\mathrm{cumul}} = \sqrt{
    \underset{x_k \in X}{\mathbb{E}} \Bigl\{ [\mathrm{dif_{i,j}}(x_k)]^2 \Bigr\} }
\]
\noindent is obtained using $100\%$ of the data objects in $x_k \in X$.
Here we denote $\delta^r_{i,j}$ as being obtained using $r$\% of $X$ ($0 < r < 100$).
In this work, we set $r=90$.
\[
    \phi_{\text{ssty}} = \frac{2}{n(n+1)} \sum\nolimits^{n}_{i=1} \sum\nolimits^{i-1}_{j=0} (\delta_{i,j} - \delta^{r}_{i,j})^2
\]

\paragraph{6. Horizontal Gradient (Mean).} The mean value of the gradient values of the variance matrices in the horizontal direction.
\begin{align*}
    e_{i,j}^{hg} &= \delta_{i, j-1} - \delta_{i,j}
    \quad i=1, 2, \ldots, n; j=1, 2, \ldots, i
\end{align*}
\[
    \phi_{\text{hg-mean}} = \frac{2}{n(n+1)} \sum\nolimits_{i=1}^n
     \sum\nolimits_{j=1}^{i} e_{i,j}^{hg}
\]
\noindent when computing all gradient measurements, the primary diagonal of the variance matrix is filled in using the average value of its neighbour(s).

\paragraph{7. Horizontal Gradient (Standard deviation).} The standard deviation value of the gradient values of the variance matrices in the horizontal direction.
\[
    \phi_{\text{hg-std}}^2 = \frac{2}{n(n+1)} \sum\nolimits_{i=1}^n
     \sum\nolimits_{j=1}^{i} (e_{i,j}^{hg} - \phi_{\text{hg-mean}})^2
\]

\paragraph{8. Horizontal Gradient (row-based standard deviation).} The averaged standard deviation value of each column (valid elements) of the gradient map $[e^{hg}_{i,j}]$.
\begin{align*}
    \mu_i &= \frac{1}{i} \sum^{i}_{j=1} e_{i,j}^{hg}
    \quad i=1, 2, \ldots, n\\
    \phi_{\text{hg-rstd}} &= \frac{1}{n} \sum\nolimits_{i=1}^n
    \sqrt{ \frac{1}{i} \sum\nolimits_{j=1}^{i} (e_{i,j}^{hg} - \mu_i)^2 }
\end{align*}

\paragraph{9. Vertical Gradient (Mean).} The mean value of the gradient values of the variance matrices in the vertical direction.

\begin{align*}
    e_{i,j}^{vg} &= \delta_{i+1, j} - \delta_{i,j}
    \quad j=0, 1, \ldots, n-1; i= j, \ldots, n-1
\end{align*}
\[
    \phi_{\text{vg-mean}} = \frac{2}{n(n+1)} \sum\nolimits_{i=0}^{n-1}
     \sum\nolimits_{j=0}^{i} e_{i,j}^{vg}
\]

\paragraph{10. Vertical Gradient (Standard deviation).} The standard deviation value of the gradient values of the variance matrices in the vertical direction.
\[
    \phi_{\text{vg-std}}^2 = \frac{2}{n(n+1)} \sum\nolimits_{i=0}^{n-1}
     \sum\nolimits_{j=0}^{i} (e_{i,j}^{vg} - \phi_{\text{vg-mean}})^2
\]

\paragraph{11. Vertical Gradient (Column-based standard deviation).} The averaged standard deviation value of each row (valid elements) of the gradient map $[e^{vg}_{i,j}]$.
\begin{align*}
    \mu_j &= \frac{1}{n-j} \sum^{n-1}_{i=j} e_{i,j}^{vg}
    \quad j=0, 1, \ldots, n-1\\
    \phi_{\text{vg-cstd}} &= \frac{1}{n} \sum\nolimits_{j=0}^{n-1}
    \sqrt{ \frac{1}{n-j} \sum\nolimits_{i=j}^{n-1} (e_{i,j}^{vg} - \mu_j)^2 }
\end{align*}

\paragraph{12. Diagonal Gradient (Mean).} The mean value of the gradient values of the variance matrices in the diagonal direction.
\begin{align*}
    e_{i,j}^{dg} &= \delta_{i+1, j-1} - \delta_{i,j}
    \quad i=1, 2, \ldots, n-1; j= 1, 2, \ldots, i
\end{align*}
\[
    \phi_{\text{dg-mean}} = \frac{2}{n(n-1)} \sum\nolimits^{n-1}_{i=1} \sum\nolimits^i_{j=1} e_{i,j}^{dg}
\]

\paragraph{13. Diagonal Gradient (Standard deviation).} The standard deviation value of the gradient values of the variance matrices in the diagonal direction.

\begin{align*}
    \phi_{\text{dg-std}}^2 &= \frac{2}{n(n-1)} \sum\nolimits_{i=1}^{n-1}
    \sum\nolimits_{j=1}^{i} (e_{i,j}^{dg} - \phi_{\text{dg-mean}})^2
\end{align*}

\paragraph{14. Overall Gradient.} The averaged ratio of the mean and the standard deviation value of the gradient in horizontal/vertical/diagonal direction.
\[
    \phi_{\text{G-overall}} = (\frac{\phi_{\text{hg-mean}}}{\phi_{\text{hg-std}}}+\frac{\phi_{\text{vg-mean}}}{\phi_{\text{vg-std}}}+\frac{\phi_{\text{dg-mean}}}{\phi_{\text{dg-std}}})/3
\]

\paragraph{15. Discontinuity.} The discontinuity (flatness) value of the variance matrix.
\begin{align*}
    \mu_r = \sum\nolimits^{n-r}_{j=0}\frac{\delta_{j+r,j}}{n-r+1}
    \quad r=1, 2, \ldots, n-1
\end{align*}
\[
    \phi_{\text{dctny}} = \sum\nolimits^{n-1}_{r=1} \sum\nolimits^{n-r}_{j=0} (\delta_{j+r,j}-\mu_r)^2 / \phi_{\text{mean}}
\]

\paragraph{16. Asymmetry.} The asymmetry (about the second diagonal) value of the variance matrix.
\[
    \phi_{\text{asymm}} = \sum\nolimits^{n}_{i=1} \sum\nolimits^{i-1}_{j=0} \left | \delta_{i,j} - \delta_{n-j, n-i} \right | / \phi_{\text{mean}}
\]

\noindent\textbf{Proposition} \textbf{1} When the $\mathrm{dif}_{i,j}()$ function can be written as $f_i()-f_j()$, e.g., \emph{max} or \emph{mean}, the square value mean $\phi_{\text{svm}}$ is proportional to the difference between the variance of $f(y)$, where $y\in T(X)$, and the covariance of $f(z_i)$ and $f(z_j)$, where the latent variables $z_i=t_i(x)$ and $z_j=t_j(x)$ (of the same data object $x\in X$) are identically and dependently distributed. $T(X)$ consists of all transformed data objects.
\[
  \begin{array}{l}
    Proof\text{: Square Mean of the Matrix}: \frac{1}{2\tau^2}\sum_i \sum_j\delta_{i,j}^2 \\
    =\frac{1}{2\tau^2}\sum_i \sum_j \dE_{x\in X}[f_i^2(x)+f_j^2(x)-2f_i(x)f_j(x)] \\
    =\underbrace{\frac{1}{\tau}\sum_i\dE_{x\in X}[f_i^2(x)]}_{ \cong \E_{y\in T(X)}[f^2(y)]}
    -\underbrace{\frac{1}{\tau^2}\sum_i \sum_j 
    \dE_{x\in X}[f_i(x)f_j(x)]}_{\cong \E_{z_i,z_j\in T(X)}[f(z_i)f(z_j)]} \\
    \cong \dE_{T(X)}[f^2(y)]- \dE_{T(X)}[f(z_i)]*\dE_{T(X)}[f(z_j)] \\
    \quad\quad\qquad\qquad - \displaystyle\mathop{\mathbb{COV}}_{T(X)}[f(z_i), f(z_j)]
    \\
    =\dE_{T(X)}[f^2(y)]-\dE_{T(X)}[f(y)]^2 - \displaystyle\mathop{\mathbb{COV}}_{T(X)}[f(z_i), f(z_j)]\\
    =\displaystyle\mathop{\mathbb{VAR}}_{T(X)}[f(y)] - \displaystyle\mathop{\mathbb{COV}}_{T(X)}[f(z_i), f(z_j)]
  \end{array}
\]
\noindent where we let $\tau=n+1$, and the expected value of $f(y)$, $f(z_i)$, and $f(z_j)$ are the same. $\mathbb{COV}$ stands for covariance and $\mathbb{VAR}$ stands for variance.

Table \ref{table:correlation} shows the correlation between the accuracy evaluation \cite{fuzzaugmentation} and our measurements. At the confidence score level (CONF), $\phi_{svm}$, $\phi_{dctny}$ and $\phi_{\text{G-overall}}$ have a strong correlation to robust accuracy.
\begin{table}[ht]
\caption{Correlation between measurements and Accuracy}
\begin{adjustbox}{width=.5\textwidth, center}
\begin{tabular}{ccccccc}
 & \multicolumn{2}{c}{\textbf{Square Mean}} & \multicolumn{2}{c}{\textbf{Discontinuity}} & \multicolumn{2}{c}{\textbf{Gradient Overall}} \\ \hline
          & CONF           & CONV-1  & CONF  & CONV-1  & CONF  & CONV-1  \\
RobustAcc & \textbf{-0.63} & -0.10 & -0.53 & -0.34 & -0.47 & -0.11 \\
Accuracy  & -0.12          & -0.03 & -0.06 & -0.11 & -0.15 & -0.21 \\ \hline
\end{tabular}
\label{table:correlation}
\end{adjustbox}
\end{table}

\begin{figure}[t]
    \centering
    \includegraphics[width=80mm,height=54mm]{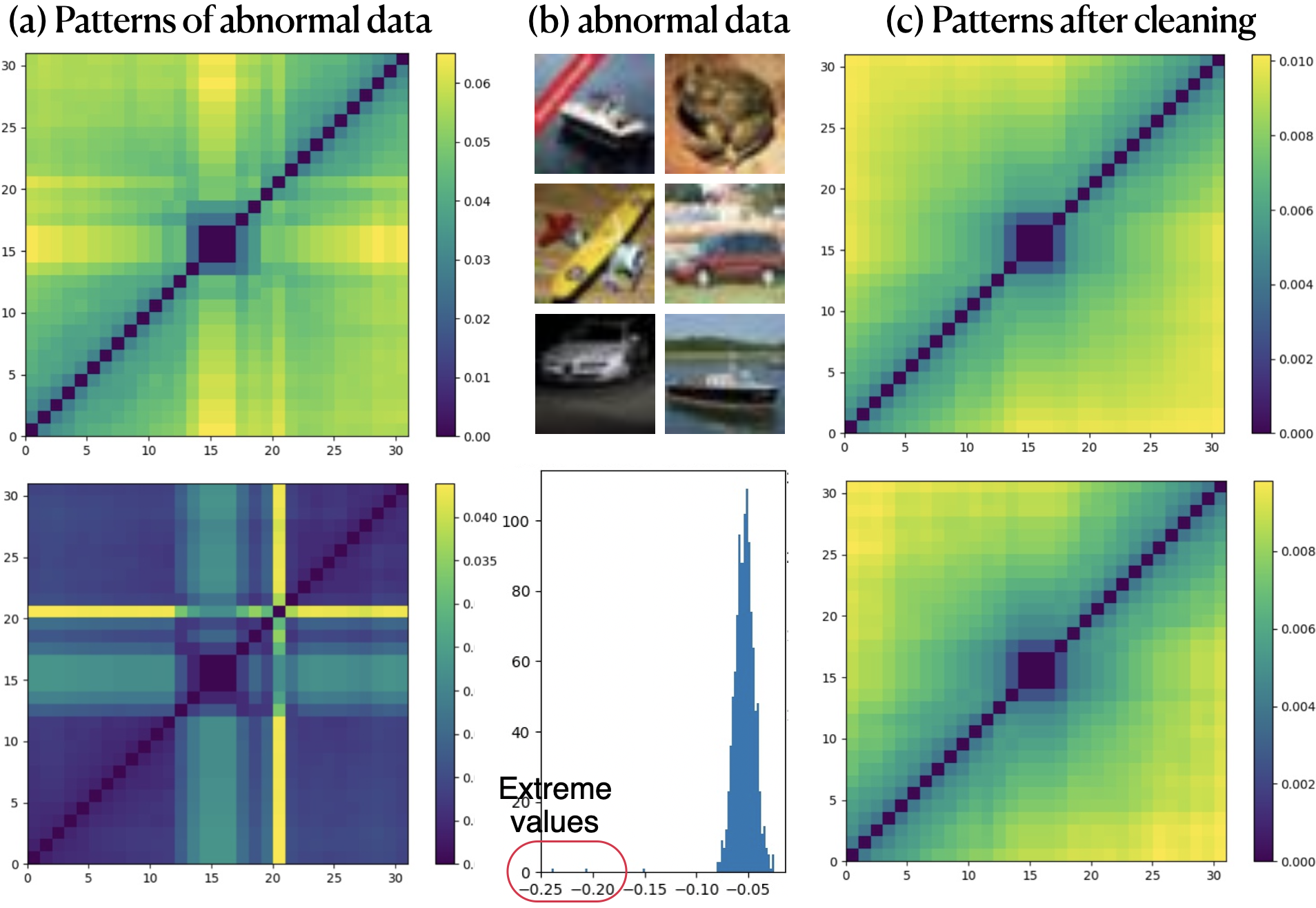}
    \caption{Rotation variance matrices \emph{mean@CONF-1} from a VGG13bn trained on CIFAR10 (model ID: 1). (a) abnormal data causes the variance matrix to be ``jagged". (b) Abnormal data lead to extreme activation values which are far away from the distribution. (c) After cleaning, the patterns become normal.}
    \label{fig:abnormaldata}
\end{figure}

\begin{table*}[t]
\caption{Metadata, accuracy and robust accuracy of the twelve CNNs (VGG13bn) shown in Fig \ref{fig:ml4ml_exp}.}
\begin{adjustbox}{max width=\textwidth, center}
\begin{tabular}{|c|c|c|c|c|c|c|c|c|c|}\hline
\diagbox[width=18mm]{Model}{Params} & Accuracy & Robust Accuracy & Learning rate & Epoch & Batch size & Pre-trained & Augmentation & FGSM trained & Anomaly \\ \hline
1 & 89.4\% & 66.3\% & $e^-5$ & 30 & 8 & \cmark & [-15$^{\circ}$, 15$^{\circ}$] & \xmark & Extreme values \\ \hline

2 & 88.0\% & 68.0\% & $e^-5$ & 30 & 8 & \cmark & [-45$^{\circ}$, 45$^{\circ}$] & \xmark & \xmark \\ \hline

8 & 89.0\% & 67.6\% & $e^-5$ & 15 & 8 & \cmark & [-20$^{\circ}$, 20$^{\circ}$] & \xmark & \xmark \\ \hline

12 & 87.8\% & 65.4\% & $e^-4$ & 3 & 8 & \cmark & [-15$^{\circ}$, 15$^{\circ}$] & \xmark & \xmark \\ \hline

13 & 89.4\% & 67.4\% & $e^-5$ & 30 & 8 & \cmark & [-15$^{\circ}$, 15$^{\circ}$] & \xmark & \xmark \\ \hline

17 & 89.1\% & 67.7\% & $e^-5$ & 2000 & 16384 & \cmark & [-10$^{\circ}$, 10$^{\circ}$] & \xmark & \xmark \\ \hline

35 & 88.0\% & 67.2\% & $e^-4$ & 50 & 1024 & \cmark & [-15$^{\circ}$, 15$^{\circ}$] & \xmark & Smaller training set \\ \hline

61 & 88.1\% & 64.3\% & $e^-4$ & 50 & 1024 & \cmark & [-5$^{\circ}$, 5$^{\circ}$] & \cmark & Data leakage \\ \hline

73 & 91.5\% & 67.5\% & $e^-4$ & 50 & 1024 & \cmark & [-5$^{\circ}$, 5$^{\circ}$] & \xmark & Impaired labels \\ \hline

85 & 89.8\% & 65.3\% & $e^-4$ & 500 & 1024 & \cmark & [-5$^{\circ}$, 5$^{\circ}$] & \xmark & Noisy data \\ \hline

96 & 85.8\% & 65.5\% & $e^-4$ & 50 & 1024 & \cmark & [-90$^{\circ}$, 90$^{\circ}$] & \xmark & \xmark \\ \hline

264 & 93.5\% & 66.3\% & $e^-4$ & 500 & 1024 & \cmark & [-10\%, +10\%] Size & \cmark & Data leakage \\ \hline
\multicolumn{8}{l}{Robust accuracy was evaluated on [-15$^{\circ}$, 15$^{\circ}$]. Accuracy was evaluated on testing set of CIFAR10.}
\end{tabular}
\end{adjustbox}
\label{table:12background}
\end{table*}

\section{Labelling Convention}
\label{sec:label_convention}
For each model, we provide the following hyper-parameters:
\begin{itemize}
    \item CNN structures (VGG13bn and CNN5).
    \item Weight initialisation: (\emph{default random initialisation}, or pre-trained on ImageNet\cite{imagenet}), weights are loaded by Pytorch \cite{pytorch}.
    \item Databases the CNNs are trained on: (CIFAR10, or MNIST).
    \item Different pre-processing: 1). Normalise pixel values to [-0.5, 0.5], 2) Normalise pixel values to [0, 1].
    \item Learning rate ($e^{-2}, e^{-3}, e^{-4}, e^{-5}$).
    \item Number of epochs (varying from 1 epoch to 2000 epochs).
    \item Batch size (4, 8, 1024 and 16384).
    \item Optimizer (e.g., Adam\cite{adam} or stochastic gradient descent(SGD)).
    \item Some models include anomalies (noisy data, data leakage etc).
    \item Whether the models are trained using gradient-based adversarial attacks (FGSM\cite{advattack}).
    \item Models trained on CPU or GPU.
    \item Different types of the transformation set $T$: \emph{rotation}, \emph{scaling} and \emph{brightness}.
    \item Different range $[-\alpha, \alpha]$ of the transformation set $T$: e.g., $\alpha = 5^\circ, 10^\circ$, etc.
\end{itemize}
Additionally, we also provide the loss functions curve on both training and testing set. The assessment labels are provided by three professional researchers/scientists who have minimum seven-year experience in both academia and industry. Note that, the labels are not the ``ground truth", since there is no standard definition about levels of invariance. Instead, they are a kind of professional annotation, which is commonly adopted for complex annotation tasks, e.g., linguistic labelling \cite{toxic}, and lying/emotion detection annotation \cite{actionunits}.

Here we describe our general guideline for labelling. Generally, we consider models as ``invariant" if the models satisfy the following conditions: 
\begin{itemize}
\item (1) they are trained with sufficient augmentation, i.e., covering the target testing interval. In our case, at least -15 to 15 for rotation, and $\times$0.7 to $\times$1.3 for scaling or brightness.
\item (2) they do not obviously underfit or overfit according to the loss function curve.
\item (3) all the concerned variance matrices are considered normal (no large values, ``cross" or ``jagged").
\item (4) during training, no anomaly that would cast doubts about the invariance quality of the ML model being labelled. For example, when 50\% of the training data objects were labelled randomly, the trained models should be labelled as ``variant".
\end{itemize}

For anomalies, we consider a CNN abnormal if we manually introduce any of the following artefacts while training:
\begin{itemize}
\item \textbf{Noisy data}: we add some meaningless random noise to our training set and label them randomly.
\item \textbf{Data leakage}: we train our models using both training set and part of testing set.
\item \textbf{Impaired label}: we manually and randomly place a wrong label to some training images.
\item \textbf{Others}: for example, no shuffling, or only use a smaller portion of the training data, or remove some intervals for augmentation, e.g., no augmentation from 0$^{\circ}$ to 5$^{\circ}$.
\end{itemize}

In addition to the anomalies above, we also found extreme values and numerical issues for VGG13bn that trained with a small batch size, e.g., 8 examples per batch. In this case, the distribution of the $\mathrm{dif}()$ function would not be sub-gaussian, therefore this leads to abrupt changes on the variance matrices, as shown in Figure \ref{fig:abnormaldata}. After those data objects that cause the extreme values have been removed, the variance matrices appear to be normal and smooth. Note that, for those CNNs with extreme values or numerical issues, they are able to achieve satisfactory classification rate (e.g., $\geqslant$ 85\% on CIFAR), and all of the abnormal data objects (which cause the issue) are correctly classified with a low confidence score (around 0.23 on average).

In order to investigate if one testing position attracts more attention than the others, we ask our three annotators the following questions:
\begin{itemize}
    \item Q1: Whether \emph{CONV-1} and \emph{CONV-2} should be checked.
    \item Q2: Whether the coder(s) always check \emph{CONV-1} and \emph{CONV-2} when labeling.
\end{itemize}
Coder 1: \emph{CONV-1} is strongly related to the feature vector(s) a model can provide for transfer learning / metric learning / self-supervised learning. Coder 2: activations at \emph{CONV-1} are often semantic features which should be considered when conducting invariance testing. Coder 3: while \emph{CONF} is often informative for invariance evaluation, \emph{CONV-1} should still be checked to ensure the invariance qualities do not only come from the final linear layer(s). All the three coders always checked \emph{CONF}. Coder 1 and Coder 2 always checked \emph{CONV-1}, while Coder 3 checked \emph{CONV-1} when they had doubts in the provided hyper-parameters or loss function curve but could not make judgement depending on variance matrices generated at \emph{CONF}. \emph{CONV-2} was checked by the three coders only when they could not make a decision after they checked variance matrices generated at \emph{CONF} and \emph{CONV-1}.


\begin{table*}[t]
\caption{Accuracy of ML4ML assessors $\mathbf{Q}(\mathbf{M}, X)$ on different model-repository partitions. Training set of the model-repository: measurements generated using half of the testing suite $X$. Testing set of the model-repository: measurements generated using the other half of the testing suite $X$.}
\begin{adjustbox}{width=164mm, center}
\begin{tabular}{|c|c|c|c|c|}
\hline
\multirow{2}{*}{\diagbox[width=50mm]{Classifier/Regressor}{Transformation}}& \multicolumn{2}{c|}{Rotation}      & Brightness         & Scaling            \\ \cline{2-5} 
                  & \emph{VGG13bn on CIFAR10} & \emph{CNN5 on MNIST} & \emph{VGG13bn on CIFAR10} & \emph{VGG13bn on CIFAR10} \\ \hline
Decision Tree     & 82.60\%$\pm$8.06\% & 81.93\%$\pm$3.93\% & 85.86\%$\pm$4.53\% & 89.66\%$\pm$3.00\% \\ \hline
Random Forest     & 91.47\%$\pm$2.13\% & \textbf{87.13\%$\pm$0.46\%} & \textbf{90.66\%$\pm$1.33\%} & \textbf{94.20\%$\pm$1.13\%} \\ \hline
AdaBoost          & \textbf{91.67\%$\pm$3.66\%} & 86.20\%$\pm$2.46\% & 88.73\%$\pm$2.73\% & 92.20\%$\pm$2.86\% \\ \hline
Linear Regression & 74.13\%$\pm$8.53\% & 67.00\%$\pm$4.33\% & 73.46\%$\pm$4.53\% & 75.73\%$\pm$6.93\% \\ \hline
\emph{*Baseline (Gao et al 2020)} & \emph{76.67}\% & \emph{83.33}\% & \emph{79.33}\% & \emph{78.67}\% \\ \hline
\multicolumn{5}{l}{\emph{All the results are averaged on ten repeated experiments. For each experiment, the standard 3-fold cross evaluation is used.}}
\end{tabular}
\end{adjustbox}
\label{table:ml4ml_results_ablation}
\end{table*}

\begin{table*}[t]
\caption{Accuracy of ML4ML assessors on different model-repository partitions. Variance matrices are generated using random proportion of the testing suite $X$ for both training and testing set of the model-repository.}
\begin{adjustbox}{width=164mm, center}
\begin{tabular}{|c|c|c|c|c|}
\hline
\multirow{2}{*}{\diagbox[width=50mm]{Classifier/Regressor}{Transformation}}& \multicolumn{2}{c|}{Rotation}      & Brightness         & Scaling            \\ \cline{2-5} 
                  & \emph{VGG13bn on CIFAR10} & \emph{CNN5 on MNIST} & \emph{VGG13bn on CIFAR10} & \emph{VGG13bn on CIFAR10} \\ \hline
Decision Tree     & 82.60\%$\pm$3.93\% & 82.67\%$\pm$2.00\% & 84.40\%$\pm$2.40\% & 88.13\%$\pm$2.80\% \\ \hline
Random Forest     & \textbf{90.73\%$\pm$1.19\%} & \textbf{87.33\%$\pm$1.66\%} & 90.46\%$\pm$1.80\% & \textbf{94.13\%$\pm$1.20\%} \\ \hline
AdaBoost          & 86.86\%$\pm$2.86\% & 86.33\%$\pm$2.13\% & \textbf{90.66\%$\pm$3.26\%} & 93.39\%$\pm$1.93\% \\ \hline
Linear Regression & 66.60\%$\pm$7.13\% & 74.60\%$\pm$5.93\% & 67.59\%$\pm$2.93\% & 68.66\%$\pm$4.66\% \\ \hline
\emph{*Baseline (Gao et al 2020)} & \emph{76.67}\% & \emph{83.33}\% & \emph{79.33}\% & \emph{78.67}\% \\ \hline
\multicolumn{5}{l}{\emph{All the results are averaged on ten repeated experiments. For each experiment, the standard 3-fold cross evaluation is used.}}
\end{tabular}
\end{adjustbox}
\label{table:ml4ml_results_ablation2}
\end{table*}

\section{Models Used for Training}
\label{sec:metadata}
We provide a model-repository with 600 models. The metadata (including all hyper-parameters) for the 600 models can be found in our metadata appendix. In order to train the ML4ML assessors, we select four ``balanced" partitions, each of which is used for training and testing the workflow $W_1\sim W_4$ introduced in Section \ref{sec:experiment}.

When testing the ML4ML assessors, although we use the standard cross-validation protocol, we also separate a ``hold-out testing set" for each partitions as shown in Table \ref{table:partitions}.
More specifically, partition 1 starts from model id: ``mid: 1" to ``mid: 100", and ``mid: t1" to ``mid: t50". When using cross-validation, the ``hold-out" testing set, i.e., ``mid: t1" to ``mid: t50", is always treated as a single fold, while the rest 100 models are randomly divided into two other folds. Note that, when training the ML4ML assessors, we did not tune any hyper-parameters, e.g., the depth for decision trees. Therefore, we did not further split the training set into training and validation set.

\begin{table}[ht]
\caption{Four partitions on the model-repository}
\begin{adjustbox}{max width=.5\textwidth, center}
\begin{tabular}{c|cccc}
\hline
           & Partition (a)      & Partition (b)      & Partition (c)      & Partition (d)     \\ \hline
Regular    & mid 1 - 100      & mid 101 - 200    & mid 201 - 300    & mid 301 - 400   \\
"Hold out" & mid t1 - t50     & mid t101 - t150  & mid t201 - t250  & mid t301 - t350 \\  \hline
\end{tabular}
\end{adjustbox}
\label{table:partitions}
\end{table}

For each partition, the statistics of the assessment labels are listed in Table \ref{table:label_stat}. Each partition is considered relatively balanced.

\begin{table}[ht]
\caption{Statistics of the assessment labels on each partition}
\begin{adjustbox}{max width=.5\textwidth, center}
\begin{tabular}{l|cccc}
\hline
\multicolumn{1}{c|}{} & Partition (a) & Partition (b) & Partition (c) & Partition (d) \\ \hline
Invariant & 66 & 78 & 56 & 86 \\
Variant & 84 & 72 & 94 & 64 \\ \hline
\end{tabular}
\end{adjustbox}
\label{table:label_stat}
\end{table}
For partition (d), we train 150 CNN5 models on MNIST. The CNN5 structure is NOT tuned. Instead, we randomly chose the hyper-parameters, e.g., the number of hidden layers, kernel size, dropout rate, etc. The CNN5 structure we use in this work is:
\begin{itemize}
    \item Input: (1, 28, 28)
    \item Conv1: kernel size (5, 5), output: 6 channels
    \item Dropout: 25\% + ReLU + MaxPooling (2, 2)
    \item Conv2: kernel size (5, 5), output: 16 channels
    \item Dropout: 25\% + ReLU + MaxPooling (2, 2)
    \item Conv3: kernel size (3, 3), output: 32 channels
    \item Dropout: 25\% + ReLU + Flatten
    \item Linear4: input 128-d, output 64-d
    \item Dropout: 25\% + ReLU
    \item Linear5: input 64-d, output 10-d
\end{itemize}
We used Pytorch \cite{pytorch} to implement the structure and the way of initialisation for the weights and biases were set to default.

\begin{table}[b]
\caption{Accuracy of ML4ML assessors on the imbalanced model repository for three different types of invariance qualities.}
\centering
\begin{adjustbox}{width=80mm, center}
\begin{tabular}{|c|c|c|c|}
\hline
                  & Rotation & Brightness & Size \\ \hline
Decision Tree     & 93.5$\pm$2.8\% & 94.4$\pm$2.0\% & 98.2$\pm$0.4\% \\ \hline
Random Forest     & 97.8$\pm$0.9\% & 97.3$\pm$0.7\% & 99.0$\pm$0.3\% \\ \hline
AdaBoost          & 96.5$\pm$1.0\% & 96.9$\pm$0.9\% & 98.9$\pm$0.7\% \\ \hline
Linear Regression & 94.8$\pm$0.8\% & 90.8$\pm$1.0\% & 96.2$\pm$0.7\% \\ \hline
baseline          & 92.2\%   & 93.1\%     & 92.8\%     \\ \hline
\end{tabular}
\end{adjustbox}
\label{table:450}
\end{table}

We also conducted experiments on the entire 450 VGG13bn in the model-repository where the data are imbalanced, i.e., there are more ``variant" models than ``invariant" ones. As shown in Table \ref{table:450}, we still observe similar results, e.g., random forest performs the best. All the results were also obtained on 10 repeated experiments. And we use the same three-fold cross-validation protocol mentioned above to carry out each experiment. The proposed framework performs $\sim$5\% better than the baseline.

\paragraph{Metadata of the models:} Table \ref{table:12background} shows all the metadata of the 12 models shown in Fig \ref{fig:ml4ml_exp}, and we provide all metadata of all the 600 models in our metadata+code appendix. Note that, when conducting machine learning testing, we should NOT rely on the metadata (they are not used to train ML4ML assessors).

\section{Mathematical Property}
\label{sec:bound_math}

\noindent\textbf{Proposition} \textbf{2} When the $\mathrm{dif}_{i,j}()$ function can be written as $f_i()-f_j()$, e.g., \emph{max} or \emph{mean}, the element $\delta_{i,j}$ of the variance matrix is strongly related to the Pearson's correlation coefficient $\rho_{f(z_i), f(z_j)}$, where the latent variables, i.e. $z_i:=t_i(x)\in T(X)$ and $z_j:=t_j(x)\in T(X)$ (of the same data object $x\in X$) are dependent and identically distributed.

$\begin{array}{l}
    Proof: \delta_{i,j}^2/2 = \dE_{T(X)} (f(z_i)-f(z_j))^2/2 \\
    = \dE_{T(X)}(f(y)^2) - \dE_{T(X)}(f(z_i)*f(z_j)) \\
    = \left [\dE_{T(X)}(f(y)^2) - \dE_{T(X)}(f(y))^2\right] -
    \\ \quad \left [\dE_{T(X)}(f(z_i)*f(z_j)) - \dE_{T(X)}(f(z_i))\dE_{T(X)}(f(z_j)) \right]
    \\
    = \mathbb{VAR}(f(y))  * (1 - \rho_{f(z_i), f(z_j)})
\end{array}$
\noindent where $\mathbb{VAR}$ stands for variance, $y\in T(X)$ is a transformed data object which has the same distribution as $z_i$/$z_j$, and the expected values of $f(y)$, $f(z_i)$ and $f(z_j)$ are the same. Moreover, $\sigma_{f(z_i)}*\sigma_{f(z_i)}=\mathbb{VAR}(f(y))$, where $\sigma$ stands for the standard deviation.

The elements of the variance matrices $\delta_{i,j}$ concentrate around the true expected value $\hat{\delta}_{i,j}$ when the $\mathrm{dif}_{i,j}()$ function is subgaussian distributed. We denote $f^{+}$ as the upper bound of the $\mathrm{dif}_{i,j}()$ function, for any $\epsilon>0$, by the Hoeffding inequality we have:
\begin{equation}
P_r(|\frac{1}{m}\sum_{k=1}^m\delta_{i,j}^2(x_k)-\hat{\delta}_{i,j}^{\;2}(x_k)|\geqslant\epsilon)\leqslant 2e^{-\frac{2*m*\epsilon^2}{{f^+}^2}}
\label{hoeffding}
\end{equation}
Therefore, when $m\gg {f^+}^2$, $\delta_{i,j}$ should not be significantly affected by the size of $X$. In Fig. \ref{fig:rotate_proportion} to Fig. \ref{fig:size_proportion}, we empirically show that our variance matrices are not heavily affected by the size of $X$, unless the distribution of $\mathrm{dif}_{i,j}()$ function is not subgaussian, e.g., Fig \ref{fig:abnormaldata} which is caused by anomalies such as numerical issues or extreme values.

\section{Ablation study}
\label{sec:ablation}
\begin{figure*}[ht]
\includegraphics[width=182mm]{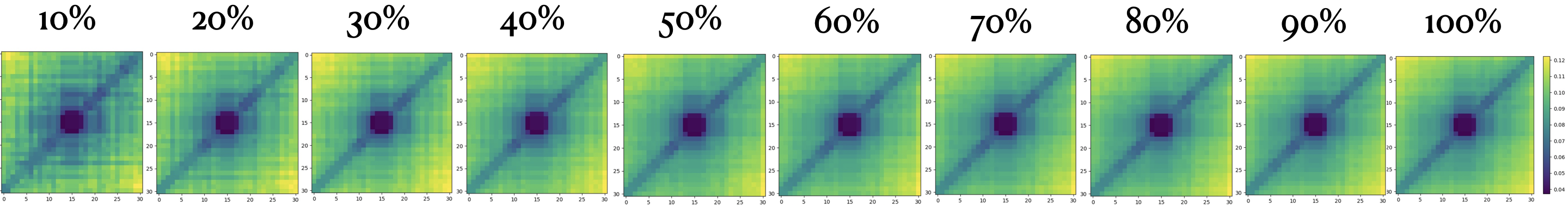}
\includegraphics[width=182mm]{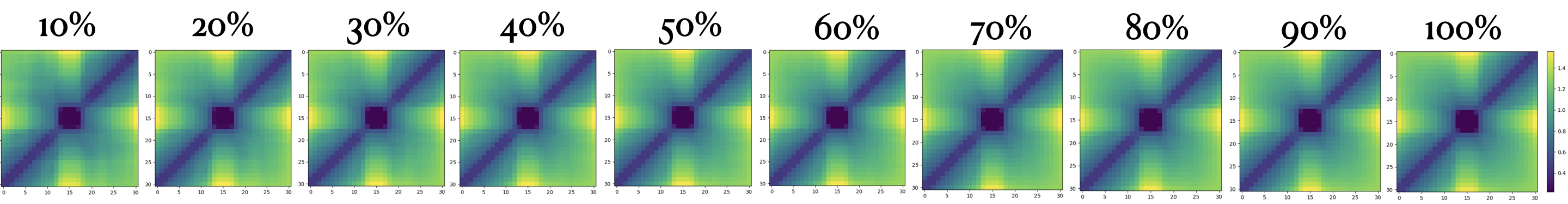}
\includegraphics[width=182mm]{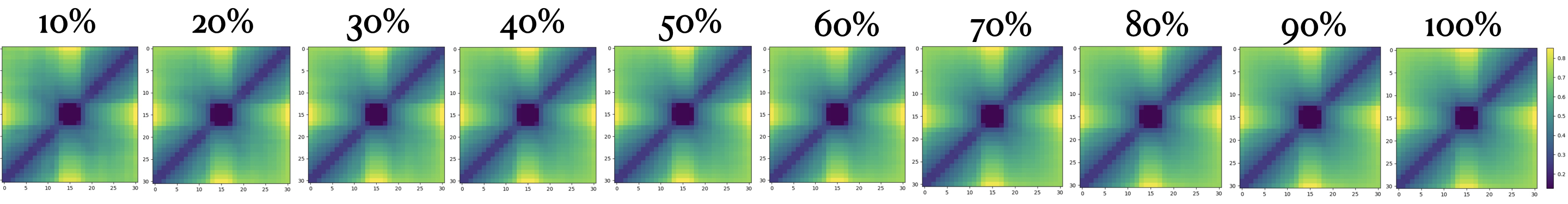}
\caption{Rotation variance matrices: (1) First row: \emph{max@CONF}; (2) Second row: \emph{max@CONF-1}, i.e., the last convolutional layer (The ninth convolutional layer for VGG13bn); (3) Third row: \emph{mean@CONF-1} of a randomly chosen VGG13bn (model id: 15) trained without anomaly. The visual patterns are obtained using a smaller proportion of the testing data. As shown in the figures, the matrices (visual patterns) are not heavily affected by the size of testing data set (i.e. the testing set of the original CIFAR10/MNIST dataset). When the proportion is greater than 30\%, we see no significant difference between the visual patterns for the first row (\emph{max@CONF}). And we see no significant difference between the visual patterns on the last two rows.}
\label{fig:rotate_proportion}
\end{figure*}

\begin{figure*}[ht]
\includegraphics[width=181.5mm]{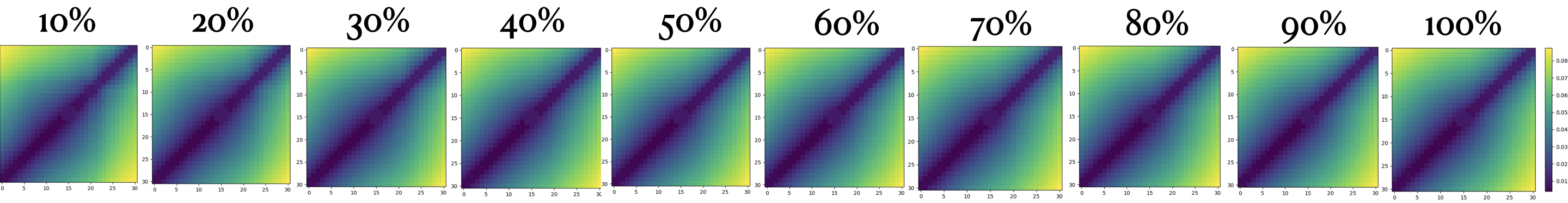}
\includegraphics[width=181.5mm]{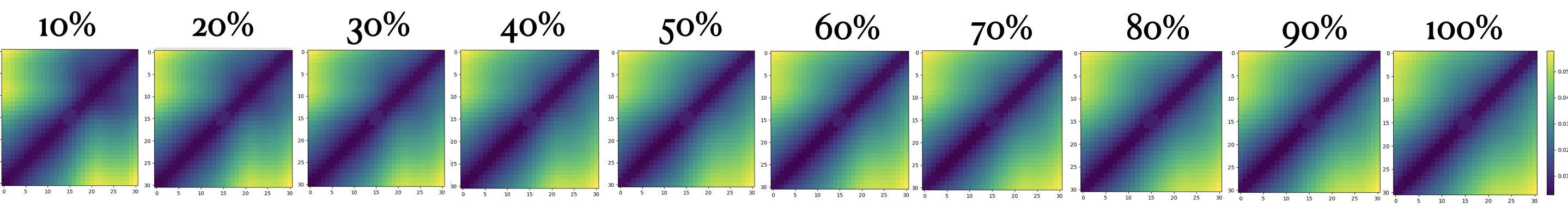}
\includegraphics[width=181.5mm]{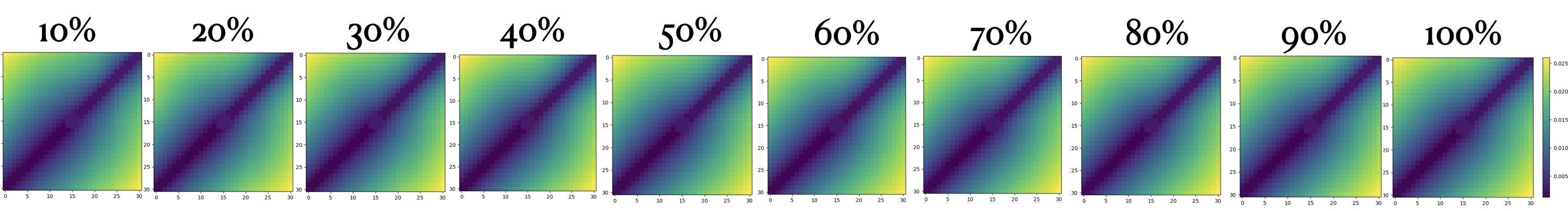}
\caption{Brightness variance matrices: (1) First row: \emph{max@CONF}; (2) Second row: \emph{max@CONF-1}, i.e., the last convolutional layer (CONV9) for VGG13bn; (3) Third row: \emph{mean@CONF-2}, i.e., the penultimate convolutional layer (CONV8) for VGG13bn, of a randomly chosen VGG13bn (model id: 102) trained without anomaly obtained using smaller proportions of the testing data. The matrices are not heavily affected by the size of testing data set (i.e. the testing set of the original CIFAR10/MNIST dataset).}
\label{fig:bright_proportion}
\end{figure*}

\begin{figure*}[ht]
\includegraphics[width=181.5mm]{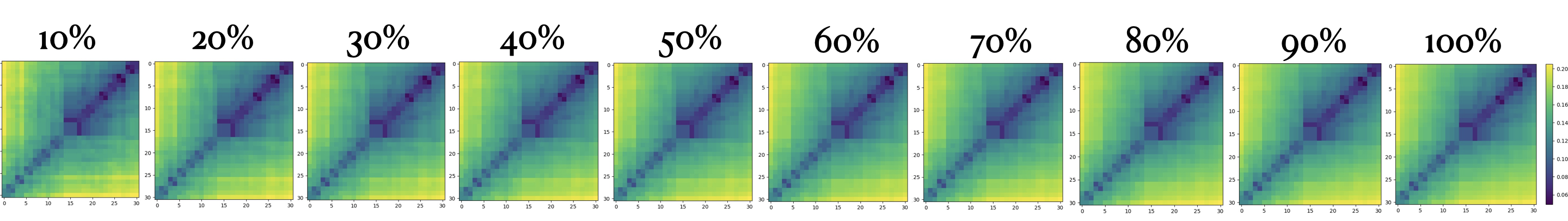}
\includegraphics[width=181.5mm]{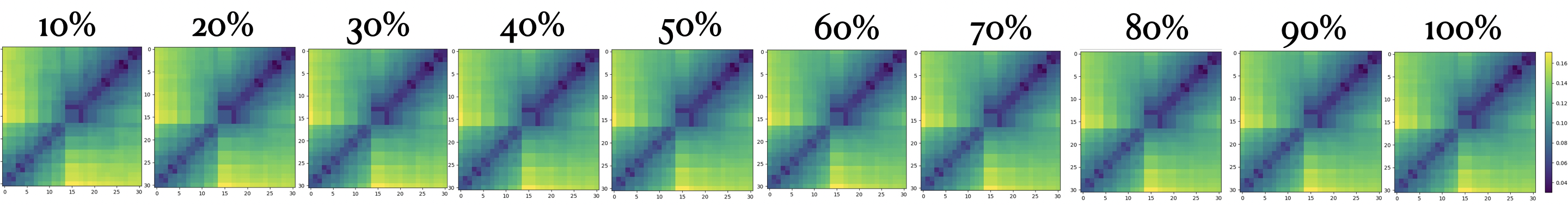}
\includegraphics[width=181.5mm]{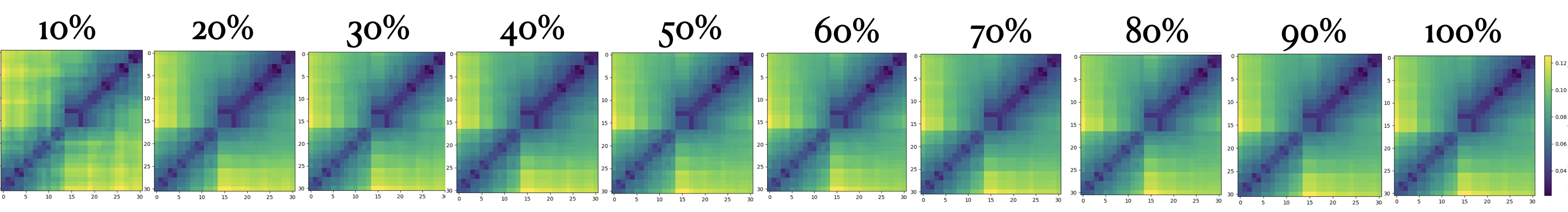}
\caption{Scaling (Size) variance matrices: (1) First row: \emph{max@CONF}; (2) Second row: \emph{max@CONF-1}, i.e., the last convolutional layer (CONV9) for VGG13bn; (3) Third row: \emph{mean@CONF-1} of a randomly chosen VGG13bn (model id: 202) trained without anomaly obtained using smaller proportions of the testing data. The matrices are not heavily affected by the size of testing data set (i.e. the testing set of the original CIFAR10/MNIST dataset).}
\label{fig:size_proportion}
\end{figure*}

\textbf{Size of testing suite.} In Fig \ref{fig:rotate_proportion} to \ref{fig:size_proportion}, we show that neither the variance matrices nor the measurements will be heavily affected by the size of the testing suite $X$.

\textbf{Splitting testing suites.} We split the testing suite $X$ into two parts, and use one part to generate variance matrices and measurements for the models on the training set of the balanced model-repository, and use the other part to generate variance matrices and measurements for the models on the testing set of the balanced model-repository. In Table \ref{table:ml4ml_results_ablation}, we show that, under this setting, our testing framework (the ML4ML assessors) can still achieve satisfactory results (over 90\% on the testing set of the balanced model-repository). Note that this setting, there is no overlapping for both the ML models, or the testing suite for the training/testing set of the model-repository.

Moreover, it is not necessary to keep the same proportion of testing suite $X$ for either training or testing the ML4ML assessors. In Table \ref{table:ml4ml_results_ablation2}, we show that when generating the measurements using random proportion of the testing suite $X$, our testing framework (the ML4ML assessors) is still able to achieve satisfactory results (over 90\% on the testing set of the balanced model-repository).


\newpage
\hbox{}
\newpage

\begin{table*}[hb]
\centering
\caption{Measurements on rotation variance matrices (\emph{max@CONF}) of a randomly chosen VGG13bn (model id: 15) trained without anomaly, and the visual patterns are obtained using a smaller proportion of the testing data. The measurements are not heavily affected by the size of testing data set (i.e. the testing set of the original CIFAR10/MNIST dataset).}

\begin{adjustbox}{width=180mm, center}
\begin{tabular}{|c|c|c|c|c|c|c|c|c|c|c|}
\hline
\diagbox[width=35mm]{Proportion}{Measurement}& 10\%  & 20\%  & 30\%  & 40\%  & 50\%  & 60\%  & 70\%  & 80\%  & 90\%  & 100\% \\ \hline
Mean          & 0.085 & 0.089 & 0.089 & 0.087 & 0.087 & 0.087 & 0.087 & 0.087 & 0.087 & 0.088 \\ \hline
Discontinuity & 1.523 & 1.418 & 1.326 & 1.364 & 1.372 & 1.283 & 1.274 & 1.278 & 1.275 & 1.258 \\ \hline
Asymmetry     & 0.074 & 0.043 & 0.034 & 0.033 & 0.032 & 0.029 & 0.031 & 0.032 & 0.029 & 0.029 \\ \hline
Gradient      & 0.730 & 0.766 & 0.780 & 0.806 & 0.826 & 0.824 & 0.836 & 0.834 & 0.834 & 0.835 \\ \hline
\end{tabular}
\end{adjustbox}

\vspace{5mm}
\caption{Measurements on rotation variance matrices (\emph{max@CONV-1}) of a randomly chosen VGG13bn (model id: 15) trained without anomaly, and the visual patterns are obtained using a smaller proportion of the testing data. The measurements are not heavily affected by the size of testing data set (i.e. the testing set of the original CIFAR10/MNIST dataset).}
\begin{adjustbox}{width=180mm, center}
\begin{tabular}{|c|c|c|c|c|c|c|c|c|c|c|}
\hline
\diagbox[width=35mm]{Proportion}{Measurement}& 10\%  & 20\%  & 30\%  & 40\%  & 50\%  & 60\%  & 70\%  & 80\%  & 90\%  & 100\% \\ \hline
Mean          & 0.955 & 0.931 & 0.926 & 0.926 & 0.938 & 0.934 & 0.934 & 0.934 & 0.934 & 0.934 \\ \hline
Discontinuity & 1.889 & 1.755 & 1.733 & 1.717 & 1.725 & 1.700 & 1.713 & 1.707 & 1.707 & 1.707 \\ \hline
Asymmetry     & 0.027 & 0.018 & 0.016 & 0.019 & 0.017 & 0.018 & 0.021 & 0.020 & 0.019 & 0.018 \\ \hline
Gradient      & 1.033 & 1.058 & 1.048 & 1.045 & 1.053 & 1.056 & 1.054 & 1.059 & 1.060 & 1.058 \\ \hline
\end{tabular}
\end{adjustbox}

\vspace{5mm}
\caption{Measurements on rotation variance matrices (\emph{mean@CONV-1}) of a randomly chosen VGG13bn (model id: 15) trained without anomaly, and the visual patterns are obtained using a smaller proportion of the testing data. The measurements are not heavily affected by the size of testing data set (i.e. the testing set of the original CIFAR10/MNIST dataset).}
\begin{adjustbox}{width=180mm, center}
\begin{tabular}{|c|c|c|c|c|c|c|c|c|c|c|}
\hline
\diagbox[width=35mm]{Proportion}{Measurement}& 10\%  & 20\%  & 30\%  & 40\%  & 50\%  & 60\%  & 70\%  & 80\%  & 90\%  & 100\% \\ \hline
Mean          & 0.521 & 0.512 & 0.508 & 0.511 & 0.517 & 0.516 & 0.517 & 0.517 & 0.516 & 0.517 \\ \hline
Discontinuity & 1.824 & 1.727 & 1.768 & 1.739 & 1.761 & 1.742 & 1.744 & 1.745 & 1.740 & 1.750 \\ \hline
Asymmetry     & 0.026 & 0.017 & 0.013 & 0.011 & 0.014 & 0.014 & 0.015 & 0.012 & 0.010 & 0.011 \\ \hline
Gradient      & 1.040 & 1.052 & 1.037 & 1.026 & 1.043 & 1.049 & 1.053 & 1.049 & 1.050 & 1.050 \\ \hline
\end{tabular}
\end{adjustbox}
\end{table*}

\begin{table*}[ht]
\centering
\caption{Measurements on brightness variance matrices (\emph{max@CONF}) of a randomly chosen VGG13bn (model id: 102) trained without anomaly obtained using smaller proportions of the testing data. The measurements are not heavily affected by the size of testing data set (i.e. the testing set of the original CIFAR10/MNIST dataset).}

\begin{adjustbox}{width=180mm, center}
\begin{tabular}{|c|c|c|c|c|c|c|c|c|c|c|}
\hline
\diagbox[width=35mm]{Proportion}{Measurement}& 10\%  & 20\%  & 30\%  & 40\%  & 50\%  & 60\%  & 70\%  & 80\%  & 90\%  & 100\% \\ \hline
Mean          & 0.041 & 0.041 & 0.040 & 0.040 & 0.040 & 0.040 & 0.040 & 0.040 & 0.040 & 0.040 \\ \hline
Discontinuity & 2.445 & 2.509 & 2.440 & 2.300 & 2.133 & 2.157 & 2.179 & 2.236 & 2.205 & 2.246 \\ \hline
Asymmetry     & 0.260 & 0.252 & 0.246 & 0.232 & 0.218 & 0.224 & 0.230 & 0.236 & 0.230 & 0.236 \\ \hline
Gradient      & 1.923 & 2.035 & 2.064 &2.050 & 2.023 & 2.057 & 2.049 & 2.061 & 2.049 & 2.051 \\ \hline
\end{tabular}
\end{adjustbox}

\vspace{8mm}
\caption{Measurements on brightness variance matrices (\emph{max@CONV-1}) of a randomly chosen VGG13bn (model id: 102) trained without anomaly obtained using smaller proportions of the testing data. The measurements are not heavily affected by the size of testing data set (i.e. the testing set of the original CIFAR10/MNIST dataset).}
\begin{adjustbox}{width=180mm, center}
\begin{tabular}{|c|c|c|c|c|c|c|c|c|c|c|}
\hline
\diagbox[width=35mm]{Proportion}{Measurement}& 10\%  & 20\%  & 30\%  & 40\%  & 50\%  & 60\%  & 70\%  & 80\%  & 90\%  & 100\% \\ \hline
Mean          & 0.045 & 0.035 & 0.031 & 0.029 & 0.029 & 0.028 & 0.027 & 0.026 & 0.027 & 0.026 \\ \hline
Discontinuity & 1.545 & 1.562 & 1.547 & 1.266 & 1.113 & 1.092 & 1.048 & 1.060 & 1.022 & 0.986 \\ \hline
Asymmetry     & 0.225 & 0.118 & 0.121 & 0.131 & 0.129 & 0.127 & 0.138 & 0.131 & 0.137 & 0.137 \\ \hline
Gradient      & 1.963 & 2.027 & 2.006 & 2.101 & 2.081 & 2.019 & 2.035 & 2.046 & 2.060 & 2.069 \\ \hline
\end{tabular}
\end{adjustbox}

\vspace{8mm}
\caption{Measurements on brightness variance matrices (\emph{mean@CONV-2}) of a randomly chosen VGG13bn (model id: 102) trained without anomaly obtained using smaller proportions of the testing data. The measurements are not heavily affected by the size of testing data set (i.e. the testing set of the original CIFAR10/MNIST dataset).}
\begin{adjustbox}{width=180mm, center}
\begin{tabular}{|c|c|c|c|c|c|c|c|c|c|c|}
\hline
\diagbox[width=35mm]{Proportion}{Measurement}& 10\%  & 20\%  & 30\%  & 40\%  & 50\%  & 60\%  & 70\%  & 80\%  & 90\%  & 100\% \\ \hline
Mean          & 0.012 & 0.012 & 0.012 & 0.012 & 0.012 & 0.012 & 0.012 & 0.012 & 0.012 & 0.012 \\ \hline
Discontinuity & 1.892 & 1.834 & 1.895 & 1.960 & 1.934 & 1.881 & 1.845 & 1.817 & 1.822 & 1.812 \\ \hline
Asymmetry     & 0.202 & 0.183 & 0.194 & 0.206 & 0.198 & 0.194 & 0.191 & 0.190 & 0.191 & 0.188 \\ \hline
Gradient      & 2.113 & 2.032 & 2.081 & 2.090 & 2.076 & 2.066 & 2.060 & 2.079 & 2.078 & 2.081 \\ \hline
\end{tabular}
\end{adjustbox}
\end{table*}

\begin{table*}[ht]
\centering
\caption{Measurements on scaling variance matrices (\emph{max@CONF}) of a randomly chosen VGG13bn (model id: 202) trained without anomaly obtained using smaller proportions of the testing data. The measurements are not heavily affected by the size of testing data set (i.e. the testing set of the original CIFAR10/MNIST dataset).}
\begin{adjustbox}{width=180mm, center}
\begin{tabular}{|c|c|c|c|c|c|c|c|c|c|c|}
\hline
\diagbox[width=35mm]{Proportion}{Measurement}& 10\%  & 20\%  & 30\%  & 40\%  & 50\%  & 60\%  & 70\%  & 80\%  & 90\%  & 100\% \\ \hline
Mean          & 0.137 & 0.139 & 0.141 & 0.139 & 0.138 & 0.138 & 0.139 & 0.138 & 0.138 & 0.139 \\ \hline
Discontinuity & 1.914 & 1.848 & 1.802 & 1.787 & 1.784 & 1.736 & 1.747 & 1.738 & 1.738 & 1.745 \\ \hline
Asymmetry     & 0.202 & 0.196 & 0.193 & 0.193 & 0.187 & 0.189 & 0.189 & 0.190 & 0.188 & 0.190 \\ \hline
Gradient      & 0.656 & 0.718 & 0.717 & 0.716 & 0.716 & 0.724 & 0.734 & 0.719 & 0.725 & 0.721 \\ \hline
\end{tabular}
\end{adjustbox}

\vspace{7mm}
\caption{Measurements on scaling variance matrices (\emph{max@CONV-1}) of a randomly chosen VGG13bn (model id: 202) trained without anomaly obtained using smaller proportions of the testing data. The measurements are not heavily affected by the size of testing data set (i.e. the testing set of the original CIFAR10/MNIST dataset).}
\begin{adjustbox}{width=180mm, center}
\begin{tabular}{|c|c|c|c|c|c|c|c|c|c|c|}
\hline
\diagbox[width=35mm]{Proportion}{Measurement}& 10\%  & 20\%  & 30\%  & 40\%  & 50\%  & 60\%  & 70\%  & 80\%  & 90\%  & 100\% \\ \hline
Mean          & 0.107 & 0.107 & 0.107 & 0.107 & 0.107 & 0.106 & 0.107 & 0.106 & 0.106 & 0.106 \\ \hline
Discontinuity & 1.941 & 1.809 & 1.751 & 1.748 & 1.746 & 1.780 & 1.757 & 1.772 & 1.785 & 1.782 \\ \hline
Asymmetry     & 0.211 & 0.205 & 0.206 & 0.205 & 0.203 & 0.203 & 0.202 & 0.202 & 0.201 & 0.201 \\ \hline
Gradient      & 0.757 & 0.768 & 0.779 & 0.763 & 0.756 & 0.747 & 0.744 & 0.751 & 0.755 & 0.754 \\ \hline
\end{tabular}
\end{adjustbox}

\vspace{7mm}
\caption{Measurements on scaling variance matrices (\emph{mean@CONV-1}) of a randomly chosen VGG13bn (model id: 202) trained without anomaly obtained using smaller proportions of the testing data. The measurements are not heavily affected by the size of testing data set (i.e. the testing set of the original CIFAR10/MNIST dataset).}
\begin{adjustbox}{width=180mm, center}
\begin{tabular}{|c|c|c|c|c|c|c|c|c|c|c|}
\hline
\diagbox[width=35mm]{Proportion}{Measurement}& 10\%  & 20\%  & 30\%  & 40\%  & 50\%  & 60\%  & 70\%  & 80\%  & 90\%  & 100\% \\ \hline
Mean          & 0.076 & 0.076 & 0.076 & 0.076 & 0.076 & 0.076 & 0.076 & 0.076 & 0.076 & 0.076 \\ \hline
Discontinuity & 1.974 & 2.184 & 2.027 & 2.011 & 2.009 & 2.017 & 1.992 & 1.987 & 2.020 & 2.030 \\ \hline
Asymmetry     & 0.215 & 0.234 & 0.216 & 0.214 & 0.214 & 0.213 & 0.214 & 0.210 & 0.214 & 0.215 \\ \hline
Gradient      & 0.808 & 0.835 & 0.872 & 0.871 & 0.863 & 0.865 & 0.858 & 0.859 & 0.861 & 0.856 \\ \hline
\end{tabular}
\end{adjustbox}
\end{table*}

\end{document}